%% file: main.tex
\documentclass[10pt,twocolumn,letterpaper]{article}

\usepackage{cvpr}              
\usepackage{algorithm}
\usepackage{algorithmic}
\usepackage{multirow}
\usepackage{makecell}

\usepackage{mathtools}
\usepackage{graphicx}
\usepackage{nicefrac}
\definecolor{cvprblue}{rgb}{0.21,0.49,0.74}
\usepackage[pagebackref,breaklinks,colorlinks,allcolors=cvprblue]{hyperref}
\newcommand{\hut}[1]{\textcolor{black}{#1}}
\newcommand{\zjn}[1]{\textcolor{black}{#1}} 
\newcommand{\yr}[1]{\textcolor{black}{#1}}
\newcommand{\yrn}[1]{\textcolor{black}{#1}}
\newcommand{\car}[1]{\textcolor{black}{#1}}

\def \pzo {\phantom{0}} 
\def\paperID{3025} 
\def\confName{CVPR}
\def\confYear{2025}

\title{Improving Autoregressive \yr{Visual} Generation with \\Cluster-Oriented Token Prediction}

\author{Teng Hu$^{1\ast}$,~~~Jiangning Zhang$^{2,3}$\thanks{Equal contribution.} ,~~~Ran Yi$^{1}$\thanks{Corresponding author.} ,~~~Jieyu Weng$^{1}$,~~~Yabiao Wang$^{3,2}$,\\Xianfang Zeng$^{3}$,~~~Zhucun Xue$^{3}$,~~~Lizhuang Ma$^{1}$\\
$^1$Shanghai Jiao Tong University,~~~$^2$Youtu Lab, Tencent,~~~$^3$Zhejiang University\\
$\{$hu-teng, ranyi, w.jerry, lzma$\}$@sjtu.edu.cn,\\
$\{$186368, yabiaowang, zzlongjuanfeng, 12432038$\}$@zju.edu.cn\\
Code: \href{https://github.com/sjtuplayer/IAR}{\textcolor{magenta}{https://github.com/sjtuplayer/IAR}}
}

\begin{document}
\maketitle

\begin{abstract}
\yr{Employing LLMs for visual generation has recently become a research focus.}
However, \yr{the existing} methods primarily \yr{transfer} the LLM architecture to \yr{visual} generation \yr{but rarely} investigat\yr{e} the \yr{fundamental} differences between language and vision. This oversight may lead to suboptimal utilization of visual generation capabilities within the LLM framework.
In this paper, we explore \yr{the characteristics of} visual \yr{embedding space} under the LLM framework and discover that the correlation between visual embeddings can help achieve more stable and robust generation results. 
\yr{We present \hut{\textbf{IAR}, an \textbf{I}mproved \textbf{A}uto\textbf{R}egressive Visual Generation Method that enhances} the training efficiency and generation quality of LLM-based visual generation models. Firstly,}
we propose a \yr{Codebook Rearrangement strategy that uses} balanced k-means clustering algorithm to rearrange the visual codebook \yr{into clusters}, ensuring high similarity among visual features within each cluster. 
Leveraging the rearranged codebook, we \yr{propose} a Cluster-oriented Cross-entropy Loss that \yr{guides} the model \yr{to correctly predict the cluster where the target token is located.} 
This approach ensures that even if the model predicts the wrong token index, there is a high probability the predicted token is located in the correct cluster, which significantly enhances the \yr{generation} quality and robustness. 
\yr{Extensive} experiments demonstrate that our IAR \yr{consistently enhances the model training efficiency and performance from 100M to 1.4B, reducing the training time by half while achieving the same FID.}
Additionally, IAR can be applied to \yr{various} LLM-based visual generation model\yr{s} and adheres to the scaling law, providing a \yr{promising} direction for future research in LLM-based visual generation.
\end{abstract}

\input{sec/1_intro}

\input{sec/2_related_work}

\input{sec/3_method}

\input{sec/4_experiment}
\input{sec/5_summary}
{
    \small
    \bibliographystyle{ieeenat_fullname}
    \bibliography{main}
}


\clearpage

\renewcommand\thefigure{A\arabic{figure}}
\renewcommand\thetable{A\arabic{table}}  
\renewcommand\theequation{A\arabic{equation}}
\setcounter{equation}{0}
\setcounter{table}{0}
\setcounter{figure}{0}
\appendix

\section{Overview}
\label{sec:rationale}
The supplementary material is composed of:
\begin{itemize}

\item Implementation details (Sec.~\ref{sec:implementation details});

\item More details on optimization relaxation in codebook rearrangement (Sec.~\ref{sec:optimization relaxation})

\item Compar\yr{ison} with VAR~\cite{VAR} (Sec.~\ref{sec:comparing with VAR})

\item  More analysis on our model (Sec.~\ref{sec:more analysis on our model});

\item  Analysis on the cluster-oriented cross-entropy loss $\mathcal{L}_{CCE}$ (Sec.~\ref{sec:analysis on LCCE});

\item  Analysis on the influence of code-rearrangement quality (Sec.~\ref{sec: analysis on code-rearrangement quality});

\item Experiments on different VQVAEs (Sec.~\ref{sec:Experiments on different VQVAE});

\item More visualization results (Sec.~\ref{sec: more visualized results});

\item 
\yr{Future work}
(Sec.~\ref{sec:further improvement}).
\end{itemize}


\section{Implementation Details}
\label{sec:implementation details}
{\bf Metrics.} We employ four metrics to evaluate the effectiveness of the models:
\begin{itemize}
    \item \textbf{Fr\'{e}chet inception distance (FID)}~\cite{fid} measures the similarity between the features of the source data and the generated data according to their mean values and covariance. A smaller FID indicates better generation ability.
    \item \textbf{Inception Score (IS)}~\cite{inception_score} measures the quality and diversity of images by computing the information entropy of the generated images. A higher IS indicates better generation quality and diversity.
    \item \textbf{Precision/Recall}~\cite{precision_and_recall} measures the class-conditional generation accuracy. A higher precision or recall indicates a better class-conditional generation performance.
\end{itemize}

\noindent {\bf Experiment settings.} We follow the experiment settings as LlamaGen~\cite{llamagen} and keep the hyperparameters consistent with it. The experiment details are shown in Tab.~\ref{tab:training settings} and Tab.~\ref{tab:inference settings}, where Tab.~\ref{tab:inference settings} is the inference settings for Tab. 2 of the main paper.

\noindent {\bf Sampling hyperparameters:}
\yr{Among the hyperparameters used in}
the inference process (Tab.~\ref{tab:inference settings}), there are several important parameters, 
\yr{whose meanings are explained in detail below:}

\textbf{(1) Classifier-free guidance:}
Classifier-Free Guidance (CFG)~\cite{cfg} is originally a sampling method to improve diffusion models by combining conditional and unconditional score estimates. Beyond diffusion models, CFG can also be applied to the autoregressive image generation process~\cite{llamagen}. Denoting the input image token sequence as $q$, our model as $\epsilon_\theta$, and the class condition as $c$, the autoregressive CFG is defined as:
\begin{equation}
    \tilde{\epsilon}_\theta(q, c) = (1 + w)\epsilon_\theta(q, c) - w\epsilon_\theta(q,\phi),
\end{equation}
where $\phi$ denotes the empty condition and $\epsilon(,)$ represents the predicted probability distribution for the next image token.



\textbf{(2) Top-K: }
Top-K sampling~\cite{topk} is a decoding strategy that selects tokens from the top \( k \) highest-probability candidates. It focuses on the most likely tokens, but the fixed \( k \) size may exclude important low-probability options.

\textbf{(3) Top-P: }
Top-P sampling~\cite{topp}, also known as nucleus sampling, selects tokens dynamically from the smallest set whose cumulative probability meets or exceeds a threshold \( p \). This approach adapts to the output distribution, balancing coherence and diversity in text generation.

\textbf{(4) Temperature: }
In large language models (LLMs), temperature~\cite{temperature,temperature2} is a hyperparameter that controls the randomness of the generated token by adjusting the sharpness of the probability distribution: lower values make the output more deterministic, while higher values increase diversity and randomness. The probability \( P_i \) for each token is calculated as:
\[
P_i = \frac{\exp{(l_i / T)}}{\sum_j \exp{(l_j / T)}},
\]
where \( l_i \) represents the predicted probability distribution, and \( T \) is the temperature.

\begin{table*}[h]
\centering
\resizebox{0.8\textwidth}{!}{
\begin{tabular}{l|cccc|cccc}
\toprule
Model                  & B&L&XL&XXL           & B&L&XL&XXL          \\ \midrule
Parameter Num          & 111M     & 343M     & 775M     & 1.4B     & 111M     & 343M     & 775M     & 1.4B     \\ \midrule
Token Num              & \multicolumn{4}{c|}{16$\times$16}                 & \multicolumn{4}{c}{24$\times$24}                 \\ \midrule

Optimizer              & \multicolumn{8}{c}{AdamW}                                                             \\
Weight decay           & \multicolumn{8}{c}{0.05}                                                              \\
Learing Rate Scheduler & \multicolumn{8}{c}{Constant}                                                          \\ \midrule
Batch Size             & 256      & 256      & 256      & 256      & 256      & 256      & 256      & 512      \\
Learning Rate          & 1E-04 & 1E-04 & 2E-04 & 2E-04 & 1E-04 & 1E-04 & 2E-04 & 2E-04 \\
GPU Num                & 8        & 8        & 8        & 8        & 8        & 8        & 16        & 32       \\
Epoch                  & 300      & 300      & 50       & 50       & 300      & 300      & 300      & 300     \\
FSDP & No & No & No & Yes& No& No& No& Yes\\
\bottomrule
\end{tabular}}
\vspace{-0.05in}
\caption{The training settings and hyperparameters used in our model.}
\vspace{-0.1in}
\label{tab:training settings}
\end{table*}

\begin{table*}[h]
\centering
\resizebox{0.8\textwidth}{!}{
\begin{tabular}{l|cccc|cccc}
\toprule
Model                  & B&L&XL&XXL           & B&L&XL&XXL          \\ \midrule
Parameter Num  \pzo\pzo\pzo\pzo\pzo        & 111M     & 343M     & 775M     & 1.4B     & 111M     & 343M     & 775M     & 1.4B     \\ \midrule
Token Num              & \multicolumn{4}{c|}{16$\times$16}                 & \multicolumn{4}{c}{24$\times$24}                 \\ \midrule
Batch Size    & \multicolumn{8}{c}{32}                                \\
Random Seed   & \multicolumn{8}{c}{0}                                 \\
Top K         & \multicolumn{8}{c}{0}                                 \\
Top P         & \multicolumn{8}{c}{1.0}                                 \\
Temperature   & \multicolumn{8}{c}{1.0}                                 \\
 \midrule
CFG           & 2.0 & 2.0 & 1.75 & 2.0 & 2.25 & 1.75 & 1.75 & 1.65 \\
\bottomrule
\end{tabular}}
\vspace{-0.05in}
\caption{The inference settings and hyperparameters used in Tab. 2 of the main paper.}
\vspace{-0.1in}
\label{tab:inference settings}
\end{table*}

\begin{table*}[t]
\centering
\resizebox{0.8\textwidth}{!}{
\begin{tabular}{c|cccc|cccc}
\toprule
\multirow{2}{*}{\makecell[c]{Classifier-free\\Guidance}} & \multicolumn{4}{c|}{VAR-d16 + \textbf{IAR}}                 & \multicolumn{4}{c}{VAR-d16}                                    \\
                     & FID$\downarrow$  & IS$\uparrow$     & Precision$\uparrow$ & Recall$\uparrow$          & FID$\downarrow$  & IS$\uparrow$     & Precision$\uparrow$ & Recall$\uparrow$          \\ \midrule
1.5 & \textbf{4.12}&\textbf{58.15}&\textbf{0.839}&\textbf{0.482}& 4.28&56.66&0.830&0.479\\
1.75&\textbf{4.07}&\textbf{60.54}&\textbf{0.857}&0.458&4.25&59.00&0.846&\textbf{0.460}\\
2.0&\textbf{4.43}&\textbf{63.11}&\textbf{0.865}&\textbf{0.435}&4.52&61.00&0.860&0.435\\
\bottomrule
\end{tabular}}
\vspace{-0.05in}
\caption{Comparing VAR-d16~\cite{VAR} with VAR+IAR on ImageNet~\cite{deng2009imagenet}. It shows that our IAR also performs well in the next-scale prediction model, validating that our method can be widely applied to various autoregressive image generation models, enhancing their generative capabilities.}
\vspace{-0.1in}
\label{tab:compare with VAR}
\end{table*}

\begin{table*}[t]
\centering
\resizebox{0.8\textwidth}{!}{
\begin{tabular}{c|cccc|cccc}
\toprule
\multirow{2}{*}{\makecell[c]{Classifier-free\\Guidance}} & \multicolumn{4}{c|}{IAR-B}                                         & \multicolumn{4}{c}{IAR-L}                                         \\ 
                     & FID$\downarrow$  & IS$\uparrow$     & Precision$\uparrow$ & Recall$\uparrow$          & FID$\downarrow$  & IS$\uparrow$     & Precision$\uparrow$ & Recall$\uparrow$          \\ \midrule
1                    & 29.70         & 43.96           & 0.566          & \textbf{0.632} & 20.56         & 62.96           & 0.595          & \textbf{0.666} \\
1.5                  & 10.69         & 103.59          & 0.732          & 0.532          & 4.39          & 178.78          & 0.778          & 0.566          \\
1.75                 & 7.43          & 135.55          & 0.783          & 0.501          & \textbf{3.18} & 234.79          & 0.824          & 0.530          \\
2                    & 6.06          & 165.22          & 0.822          & 0.454          & 3.49          & 279.09          & 0.855          & 0.499          \\
2.25                 & \textbf{5.77} & 192.45          & 0.850          & 0.421          & 4.43          & 311.08          & 0.873          & 0.466          \\
2.5                  & 6.11          & 213.76          & 0.869          & 0.381          & 5.61          & 340.18          & 0.890          & 0.425          \\
2.75                 & 6.73          & \textbf{232.35} & \textbf{0.884} & 0.360          & 6.74          & \textbf{358.48} & \textbf{0.898} & 0.401   \\
                     \bottomrule
\end{tabular}}
\vspace{-0.05in}
\caption{The Quantitative metrics on our model under different classifier-free guidance scales.}
\vspace{-0.1in}
\label{tab:different CFGs}
\end{table*}

\begin{table*}[t]
\centering
\resizebox{0.8\textwidth}{!}{
\begin{tabular}{c|c|cccc|cccc}
\toprule
\multirow{2}{*}{Model Size} & \multirow{2}{*}{Epoch} & \multicolumn{4}{c|}{LlamaGen}               & \multicolumn{4}{c}{\textbf{IAR}}                                           \\
                            &                        & FID$\downarrow$  & IS$\uparrow$     & Precision$\uparrow$ & Recall$\uparrow$          & FID$\downarrow$           & IS$\uparrow$              & Precision$\uparrow$      & Recall$\uparrow$          \\ \midrule
\multirow{4}{*}{B Version}  & 50    & 8.67 & 136.62 & 0.818     & \textbf{0.413} & \textbf{7.80} & \textbf{153.31} & \textbf{0.839} & 0.394 \\
                            &       100                 & 7.26 & 152.50 & 0.827     & 0.416          & \textbf{6.77} & \textbf{171.73} & \textbf{0.839} & \textbf{0.416} \\
                            &      200                  & 6.54 & 167.82 & 0.833     & 0.428          & \textbf{5.86} & \textbf{185.28} & \textbf{0.845} & \textbf{0.428} \\
                            &        300                & 6.09 & 182.54 & 0.845     & 0.416          & \textbf{5.77} & \textbf{192.45} & \textbf{0.850} & \textbf{0.421} \\ \midrule
\multirow{4}{*}{L Version}  & 50    & 4.25 & 191.46 & 0.819     & 0.504          & \textbf{4.35} & \textbf{197.23} & \textbf{0.819} & \textbf{0.507} \\
                            &      100                  & 3.96 & 199.96 & 0.803     & \textbf{0.532} & \textbf{3.81} & \textbf{205.63} & \textbf{0.805} & 0.528 \\
                            &       200                 & 3.33 & 219.57 & 0.804     & 0.538          & \textbf{3.31} & \textbf{225.95} & \textbf{0.814} & \textbf{0.551} \\
                            &        300                & 3.29 & 227.83 & 0.818     & \textbf{0.532} & \textbf{3.18} & \textbf{234.79} & \textbf{0.824} & 0.530  \\
                                   \bottomrule
\end{tabular}}
\caption{The Quantitative metrics on our model and LlamaGen under different training epochs. The B version employs CFG=2.25 and the L-version employs CFG=1.75.}
\label{tab:different epochs}
\end{table*}

\begin{figure*}[t]
\centering
\includegraphics[width=0.8\textwidth]{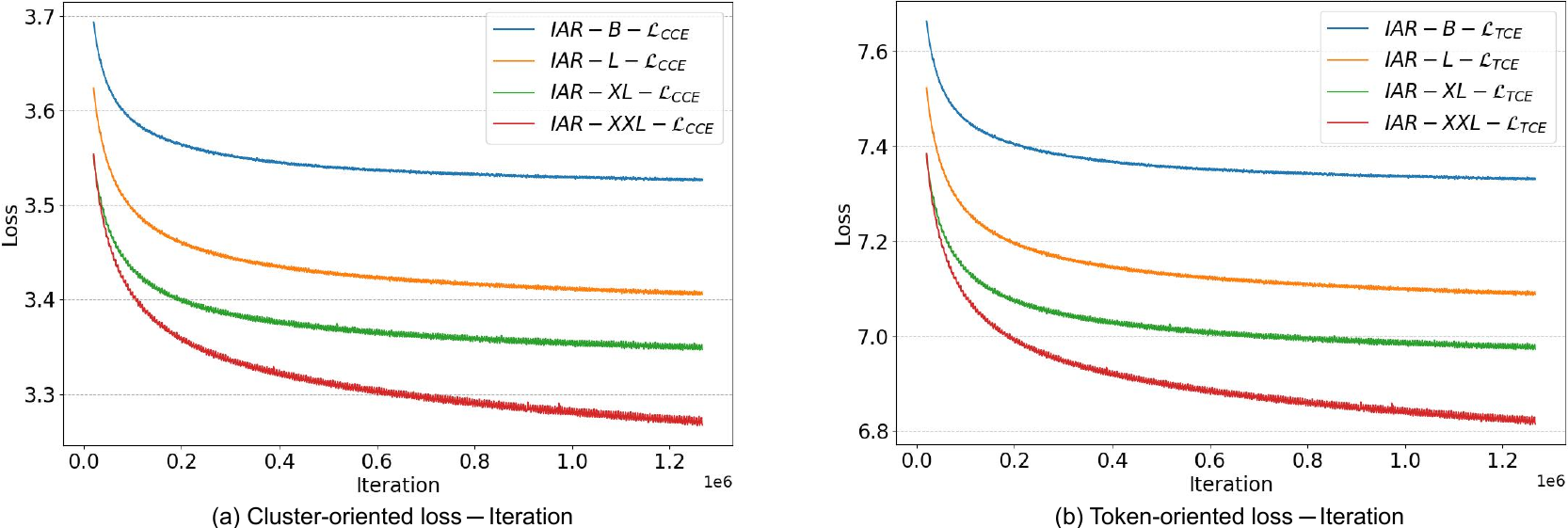}
\caption{The training loss curves for the cluster-oriented cross-entropy loss $\mathcal{L}_{CCE}$ (a) and token-oriented cross-entropy loss $\mathcal{L}_{TCE}$ (b) on $\mathbf{24\times 24}$ \textbf{image tokens}.}
\label{fig: loss curves-576}
\end{figure*}

\begin{figure*}[t]
\centering
\includegraphics[width=0.8\textwidth]{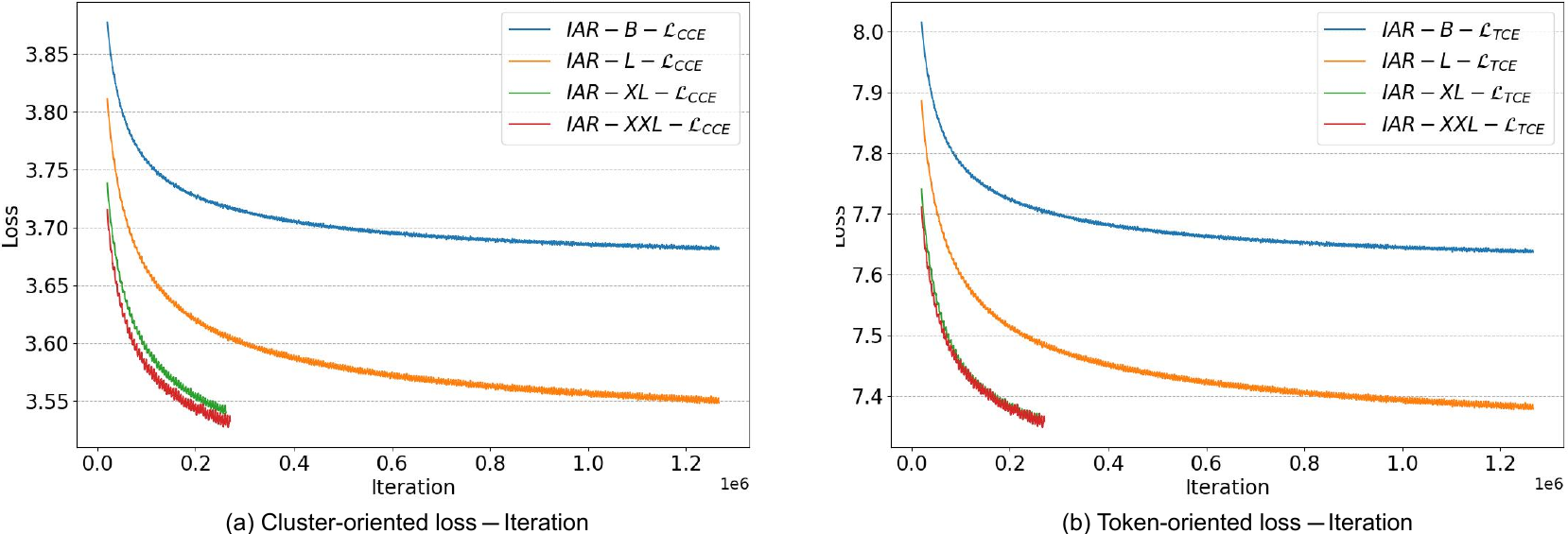}
\caption{The training loss curves for the cluster-oriented cross-entropy loss $\mathcal{L}_{CCE}$ (a) and token-oriented cross-entropy loss $\mathcal{L}_{TCE}$ (b) on $\mathbf{16\times 16}$ \textbf{image tokens}.}
\label{fig: loss-curves-384}
\end{figure*}

\section{
\yr{Complexity Analysis of Codebook Rearrangement Target}
}
\label{sec:optimization relaxation}
In Sec. 3.2 of the main paper, we aim to rearrange the codebook such that the neighboring embeddings are as close to each other. We summarize this code rearrangement problem as an optimization problem, where we aim to find a surjective mapping $M(\cdot)$ that satisfies:
\begin{equation}
    \begin{aligned}
        M=\arg\min\limits_{M} \sum_{i=\yr{1}}^{N-1} \|z_{M(i)},z_{M(i+1)}\|.
    \end{aligned}
    \label{eq:code rearrange target}
\end{equation}

After reordering each embedding $z_i$ to index $M(i)$, the sum of distances between adjacent embeddings is minimized.
And $\hat{\mathcal{Z}}=M(\mathcal{Z})$ is the rearranged codebook.

However, this optimization can be reduced to the Shortest Hamiltonian path problem, which is a classical NP-hard problem. \hut{The Shortest Hamiltonian Path Problem is a variation of the Hamiltonian Path Problem. \yr{Its goal is} to find a path that visits each vertex exactly once \yr{and} minimizes the total weight (or distance) of the path. Formally, given a weighted graph \( G = (V, E) \) with a weight function \( w: E \rightarrow \mathbb{R}^+ \), the goal is to find a Hamiltonian path $\pi^* = (\pi_1, \pi_2, \ldots, \pi_N)$ such that the sum of the weights of the edges in the path, \yr{\emph{i.e.},} $\sum_{i=1}^{N-1} w(\pi_i, \pi_{i+1})$, is minimized, which is formulated as:}

   \begin{equation}
   \pi^* = \arg\min_{\pi} \sum_{i=1}^{N-1} w(\pi_i, \pi_{i+1}).
   \label{eq: Hamiltonian path problem}
   \end{equation}

\yr{Next, w}e prove \yr{that solving the optimization problem in Eq. (\ref{eq:code rearrange target}) can be reduced to the Shortest Hamiltonian path problem in Eq. (\ref{eq: Hamiltonian path problem})}: 

\noindent\textbf{\hut{Proposition}: Solving Eq. (\ref{eq: Hamiltonian path problem}) $\leq_p$ Solving Eq. (\ref{eq:code rearrange target})}

\noindent\textbf{Proof.}

\textbf{Step 1: Construct a Complete Weighted Graph}:

   Define a complete graph $G = (V, E)$ where the vertex set $V = \{0, 1, \ldots, N-1\}$ corresponds to the $N$ embeddings in the codebook $\mathcal{Z}$. Each edge $(i, j) \in E$ is assigned a weight $w(i, j)$ \yr{that} equal\yr{s} to the distance between embeddings $z_i$ and $z_j$:
   \begin{equation}
   w(i, j) = \|z_i, z_j\|.
   \end{equation}

\textbf{Step 2: Find the Minimum Weight Hamiltonian Path}:

   Finding \yr{the shortest} Hamiltonian path $\pi = (\pi_1, \pi_2, \ldots, \pi_N)$ in $G$ aims to minimize the total weight:

   \begin{equation}
   \pi^* = \arg\min_{\pi} \sum_{i=1}^{N-1} w(\pi_i, \pi_{i+1}).
   \end{equation}

\textbf{Step 3: Mapping to the Original Problem}:

   The \yr{shortest} Hamiltonian path $\pi^*$ provides the optimal permutation $M^*$ for the \yr{optimization} problem \yr{in Eq. (\ref{eq:code rearrange target})}, where $M^*(i) = \pi_i$, for $i = 1, 2, \ldots, N$.

\textbf{In summary}, the original \yr{optimization} problem (in Eq. (\ref{eq:code rearrange target})) of finding the optimal surjective mapping $M(\cdot)$ to minimize the sum of distances between consecutive embeddings, can be reduced to finding the minimum weight Hamiltonian path in a complete weighted graph\yr{,} where the weights are given by the distances between embeddings.

Therefore, \yr{the original optimization problem is also NP-hard. And} it is necessary to relax this optimization target to a clustering problem \yr{(main paper Sec. 3.2)}, which ensures the embeddings with a cluster share high similarities.

\section{Comparing IAR+VAR with VAR}
\label{sec:comparing with VAR}

VAR~\cite{VAR} extends the next-token prediction in autoregressive image generation to next-scale prediction, enabling the model to generate images progressively from small to large scales. At each scale, VAR predicts all tokens simultaneously, significantly enhancing the inference speed of the autoregressive image generation process. Our design is independent of the model structure in autoregressive image generation, allowing us to integrate our IAR with VAR, referred to as VAR+IAR. Given that most official VAR models are trained on 256 A100 GPUs, which is highly resource-intensive, we only train the VAR-d16 model for 100 epochs on ImageNet~\cite{deng2009imagenet} and subsequently compare it with VAR+IAR.

Both models (VAR and VAR+IAR) are trained for 100 epochs with a batch size of 768, maintaining the same hyperparameters as the official VAR code. We then evaluate the trained models on different CFGs. The results, presented in Tab.~\ref{tab:compare with VAR}, demonstrate that incorporating IAR into VAR enhances the original VAR in terms of generation quality and diversity, as evidenced by improved FID and IS scores. This validates the effectiveness of our model across different autoregressive image generation frameworks, showing the great potential of our IAR in the field of autoregressive image generation.

\section{More Analysis on Our Model}
\label{sec:more analysis on our model}

{\bf Comprehensive metrics for models under different CFGs.} This section presents the comprehensive metrics (FID, IS, Precision, Recall) for the models compared in Fig. 4 (a) of the main paper. As shown in Table~\ref{tab:different CFGs}, an increase in CFG leads to higher IS and precision, while recall decreases. Unlike these three metrics, FID initially improves and then deteriorates, achieving its optimal value at an intermediate CFG. Furthermore, the optimal CFG for FID varies with model size (e.g., CFG=2.25 for IAR-B and CFG=1.75 for IAR-L).

{\bf Comprehensive metrics for models under different training epochs.} Table~\ref{tab:different epochs} presents a comparison between our IAR and LlamaGen~\cite{llamagen} over various training epochs, illustrating that our model consistently outperforms LlamaGen at all stages of training. Notably, the 200-epoch IAR-B exceeds the performance of the 300-epoch LlamaGen-B, while the 200-epoch IAR-L performs similarly to the 300-epoch LlamaGen-L, highlighting the high training efficiency of our model. (Note that all B-version models use CFG=2.25, whereas all L-version models use CFG=1.75)

{\bf Training losses for different model sizes.} We show the training loss curves for both the two losses: 1) cluster-oriented cross-entropy loss $\mathcal{L}_{CCE}$ and the token-oriented cross-entropy loss $\mathcal{L}_{TCE}$ when training on $24 \times  24$ image tokens (Fig.~\ref{fig: loss curves-576}) and $16 \times 16$ image tokens (Fig.~\ref{fig: loss-curves-384}). It can be seen that as the model size increases, both the two losses decrease faster and converge to a lower value, which aligns with the scaling law~\cite{scaling_law}. Note that since we follow the training setting of LlamaGen~\cite{llamagen}, we only train IAR-XL and IAR-XXL on $16\times 16$ image tokens for 50 epochs.

\begin{figure}[t]
\centering
\includegraphics[width=0.45\textwidth]{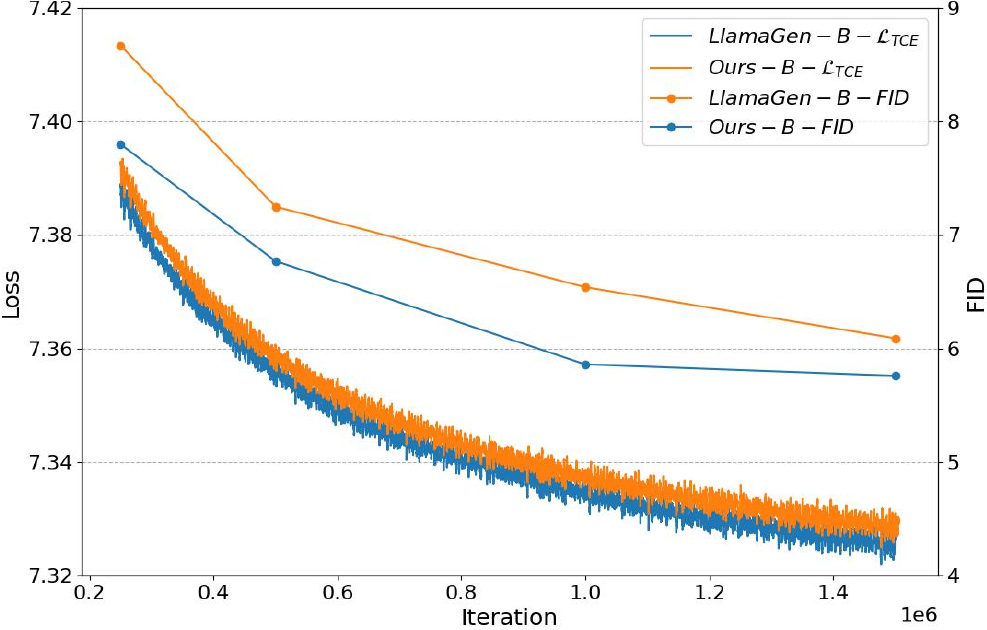}
\vspace{-0.1in}
\caption{Comparison between LlamaGen-B and ours on the \textbf{token-oriented cross-entropy loss} $\mathbf{\mathcal{L}_{TCE}}$ and the \textbf{FID} score in different training iterations. Our model has a higher $\mathcal{L}_{TCE}$ than that of LlamaGen but achieves a better FID, indicating the effectiveness of our cluster-oriented cross-entropy loss $\mathcal{L}_{CCE}$.}
\label{fig: compare L_CCE}
\vspace{-0.15in}
\end{figure}

{\bf Effectiveness of $\mathbf{\mathcal{L}_{CCE}}$.} In the main paper, we introduce the cluster-oriented cross-entropy loss $\mathcal{L}_{CCE}$, designed to enhance the model's awareness of cluster information, thereby increasing the likelihood of predicting tokens within the target cluster. It is hard to directly illustrate the effectiveness of $\mathcal{L}_{CCE}$ by its loss value directly. Therefore, we design an alternative way where we visualize the loss curves for token-oriented cross-entropy loss $\mathcal{L}_{TCE}$ and their corresponding FIDs for LlamaGen-B and our model in Fig.~\ref{fig: compare L_CCE}. The results indicate that, compared to LlamaGen, our model exhibits a higher token-oriented cross-entropy loss but achieves a superior FID. This suggests that our model has slightly lower token-oriented prediction accuracy, which is expected since the introduction of $\mathcal{L}_{CCE}$ partially diverts the original loss $\mathcal{L}_{TCE}$. Therefore, the improvement of FID can only come from our proposed cluster-oriented cross-entropy loss $\mathcal{L}_{CCE}$. Since $\mathcal{L}_{CCE}$ effectively increases the likelihood of predicting the correct cluster, combined with the embedding similarities within the cluster, it ultimately leads to the generation of images with better FID, demonstrating the efficacy of $\mathcal{L}_{CCE}$ in our model.

\begin{table*}[t]
\centering
\resizebox{0.8\textwidth}{!}{
\begin{tabular}{l|cccc|cccc}
\toprule
\multirow{2}{*}{\pzo\pzo\pzo\pzo\pzo\pzo\pzo\pzo\pzo\pzo\pzo Model}         & \multicolumn{4}{c|}{\textbf{IAR}}            & \multicolumn{4}{c}{LlamaGen~\cite{llamagen}}        \\
                               & B      & L      & XL      & XXL    & B      & L      & XL      & XXL     \\
                               \midrule
Cluster-level Accuracy (Top-1, \%) & \textbf{15.54} & \textbf{17.12} & \textbf{18.01}  & \textbf{19.02}  & 13.44          & 14.81         & 15.71         & 16.49          \\
Cluster-level Accuracy (Top-5, \%) & \textbf{41.48} & \textbf{44.68} & \textbf{46.30}  & \textbf{48.29}  & 30.37         & 33.13         & 34.87         & 36.38           \\
Token-level Accuracy (Top-1, \%)   & 2.62           & 3.17  & 3.56           & 3.88            & \textbf{2.64}  & \textbf{3.19}          & \textbf{3.59} & \textbf{3.95}  \\
Token-level Accuracy (Top-5, \%)   & 7.34           & 8.86           & 9.90           & 10.75           & \textbf{7.37} & \textbf{8.91} & \textbf{9.98} & \textbf{10.96} \\
\bottomrule
\end{tabular}}
\caption{Comparison of the token-level prediction accuracy, cluster-level prediction accuracy, and the embedding-level MSE distance between our IAR and LlamaGen.}
\label{tab:accuracy}
\end{table*}

\textbf{Token prediction accuracy.} We compute the token prediction accuracy of our model and LlamaGen~\cite{llamagen} on different model sizes ($24\times24$ tokens). Specifically, for an image token sequence $q=\{q^1,q^2,\cdots q^{576}\}$ with corresponding image embedding sequence $z_q=\{z_q^1,z_q^2,\cdots z_q^{576}\}$ and class label $c$, we enumerate $i$ from $1$ to $575$ and predict $\hat{q}^{i+1}\sim P^{i+1}=\epsilon_\theta(\hat{q}^{i+1}|c,q^1,q^2,\cdots q^{i})$ using the model $\epsilon_\theta$. We then compute the Top-1 and Top-5 accuracy $Acc^i$ between $\hat{q}^{i+1}$ and the ground truth $q^{i+1}$. The average accuracy for an image is calculated as $Acc=\frac{1}{575}\sum_{i=1}^{575} Acc^i$. 
Finally, we compute the cluster-level accuracy and token-level accuracy for all images in ImageNet~\cite{deng2009imagenet} and record the average values in Tab.~\ref{tab:accuracy}. \hut{Specifically, to compute the cluster-level accuracy for LlamaGen, we employ the balanced K-means clustering algorithm to decompose the codebook into $n$ clusters and then determine the target cluster index. We then assess whether the predicted token is located in the target cluster, thereby obtaining the cluster-level accuracy. From Tab.~\ref{tab:accuracy}, it can be seen that} our cluster-level accuracy is higher than that of LlamaGen, indicating the effectiveness of our cluster-oriented cross-entropy loss $\mathcal{L}_{CCE}$. Although our token-level accuracy is slightly lower, this is expected as the newly included loss $\mathcal{L}_{CCE}$ affects the original token-oriented cross-entropy loss $\mathcal{L}_{TCE}$, resulting in a slight decrease in token-level accuracy. However, our model still achieves better FID and IS compared to LlamaGen, further validating the effectiveness of our cluster-oriented token prediction strategy.



\section{Analysis on the Cluster-oriented Cross-entropy Loss}
 \label{sec:analysis on LCCE}


In Tab. 3 of the main paper, the model trained with only the cluster-oriented cross-entropy loss $\mathcal{L}_{CCE}$ (without codebook rearrangement) also improves generation performance (FID 6.96 vs. 7.14 for the baseline). This improvement arises because $\mathcal{L}_{CCE}$ enhances the probability of predicting the correct cluster, which is computed based on the probabilities of all tokens within the cluster. Minimizing $\mathcal{L}_{CCE}$ consequently increases the probability of the target token.
Furthermore, since $\mathcal{L}_{CCE}$ boosts the probabilities of all tokens in the target cluster, it can be viewed as a variant of label smoothing, which is known to improve generalization and model calibration in classification networks~\cite{Label-smoothing}. However, as shown in Table~\ref{tab:compare with LS}, standard label smoothing is not well-suited for autoregressive visual generation models, where token prediction accuracy is typically low (e.g., 2\%–4\% top-1 accuracy, as shown in Table~\ref{tab:accuracy}), in contrast to traditional classification tasks with significantly higher accuracy (e.g., $>70\%$). In contrast, $\mathcal{L}_{CCE}$ performs structured smoothing within specific ranges rather than uniformly smoothing all tokens, leading to improved AR generation quality. While $\mathcal{L}_{CCE}$ enhances model performance, the best results are achieved only when it is combined with our codebook rearrangement strategy. Therefore, the good performance of our IAR  is attributed to both the codebook rearrangement strategy and the cluster-oriented cross-entropy loss.

\begin{table}[t]
\centering
\renewcommand{\arraystretch}{1.1}
\resizebox{0.4\textwidth}{!}{
\begin{tabular}{l|cccc}
\toprule
Model&FID$\downarrow$ & IS$\uparrow$ & Precision$\uparrow$ & Recall$\uparrow$  \\
\midrule
LlamaGen&6.05&182.5&0.84&0.42\\
LlamaGen + IAR& \textbf{5.77}&\textbf{192.5}&\textbf{0.85}&\textbf{0.42} \\
LlamaGen + LS & 8.58&166.81&0.80&0.40\\

\bottomrule
\end{tabular}}
\caption{Comparison with the model with label smoothing, where we train LlamaGen-B with default smoothing factor 0.1 for 300 epochs.  It shows that the vanilla label smoothing is not suitable for AR generation models with low token prediction accuracy. }
\label{tab:compare with LS}
\end{table}

\section{Analysis on the Influence of Code-rearrangement Quality}
\label{sec: analysis on code-rearrangement quality}

Our IAR employs a code-rearrangement strategy to cluster similar codes, with the hypothesis that better clustering quality theoretically enhances generation performance. To validate this, we average the mean distance of each cluster as a metric to evaluate clustering quality, where a lower mean distance indicates a higher codebook rearrangement quality. We train LlamaGen-B on codebooks with varying clustering qualities (different mean distances, obtained by setting different clustering iterations in Balanced k-means Clustering) for 100 epochs to investigate how clustering quality affects performance. Results in Table~\ref{tab:fid-mean dis} demonstrate that improved clustering quality leads to better model generation performance.

\begin{table}[h]
    \centering
\renewcommand{\arraystretch}{1.1}
\resizebox{0.4\textwidth}{!}{
\begin{tabular}{c|cccc}
\toprule
Mean Dis&FID$\downarrow$ & IS$\uparrow$ & Precision$\uparrow$ & Recall$\uparrow$  \\
\midrule
0.689 &\textbf{6.77}&\textbf{171.73}&\textbf{0.84}&\textbf{0.42}\\
0.850 &6.85&170.54&0.84&0.41\\
1.239 &6.91&169.32&0.84&0.41\\
\bottomrule
\end{tabular}}
\caption{Generation performance from codebooks with different clustering quality.}
\label{tab:fid-mean dis}
\end{table}

\section{Experiments on Different VQVAEs}
\label{sec:Experiments on different VQVAE}
\car{To show the effectiveness and generalization ability of our method, we further conduct experiments on the VQVAE from VQGAN~\cite{vqgan}, where we train LlamaGen-B on it for 100 epochs. As shown in Tab.~\ref{tab:comparison on different VQVAE}, our IAR consistently improves the generation quality for the model based on VQGAN, demonstrating the effectiveness of our IAR across different VQVAEs.}

\begin{table}[t]
\centering
\renewcommand{\arraystretch}{1.1}
\centering
\resizebox{0.4\textwidth}{!}{
\begin{tabular}{lc|cccc}
\toprule
VQVAE & IAR & FID$\downarrow$ & IS$\uparrow$ & Precision$\uparrow$ & Recall$\uparrow$ \\
\midrule
LlamaGen & & 7.14 & 166.38 & 0.84 & 0.40 \\
LlamaGen & \checkmark & \textbf{6.77} & \textbf{171.73} & \textbf{0.84 }& \textbf{0.42} \\ \hline
VQGAN & & 6.90 & 176.71 & 0.84 & 0.40 \\
VQGAN & \checkmark & \textbf{6.75} & \textbf{194.64} & \textbf{0.85} & \textbf{0.40} \\ 
\bottomrule
\end{tabular}
}
\caption{Experiments on different VQVAEs. Our IAR consistently improves the performance of the model trained on different VQVQEs.}
\label{tab:comparison on different VQVAE}
\end{table}

\begin{table}[t]
\centering
\renewcommand{\arraystretch}{1.1}
\resizebox{0.4\textwidth}{!}{
\begin{tabular}{lc|ccc|c}
\toprule
VQVAE & IAR & Mean$\downarrow$ & Closest$\downarrow$ & Largest$\downarrow$ &PSNR $\uparrow$ \\
\midrule
LlamaGen & & 2.06 & 0.20 & 4.30 &20.79 \\
LlamaGen & \checkmark & \textbf{0.69} & \textbf{0.18} & \textbf{1.91} &20.79 \\ \hline
VQGAN & & 24.75 & 5e-7 & 1072.06 &20.00 \\
VQGAN & \checkmark & \textbf{8.69} & \textbf{5e-7} & \textbf{44.68} &20.00\\ 
\bottomrule
\end{tabular}}

\caption{Comparison between the original and rearranged codebooks from Llamagen and VQGAN. Our code rearrangement strategy can consistently improve the inner-cluster similarity for both the two codebooks.}
\label{tab:comparison on different VQVAE2}
\vspace{-0.18in}
\end{table}

We further compute the mean / closest / largest $\mathcal{L}_2$ distance in each cluster for the rearranged and original codebook from LlamaGen and VQGAN, to validate the effectiveness of our code rearrangement strategy. 
The results in Tab.~\ref{tab:comparison on different VQVAE2} show that our method stably decreases the three distances.  Moreover, it can be seen that the closest distance in VQGAN is $5\times10^{-7}$, indicating that some embeddings are almost the same, resulting in some embeddings being wasted (e.g., the closest distance of VQGAN (5e-7) is very low, causing the worst reconstruction PSNR (20.00)). Therefore, it motivates us that the closest distance can be used as a metric to evaluate the training of a VQVAE.

\section{More Visualization Results}
\label{sec: more visualized results}

\begin{figure*}[t]
\centering
\includegraphics[width=1.0\textwidth]{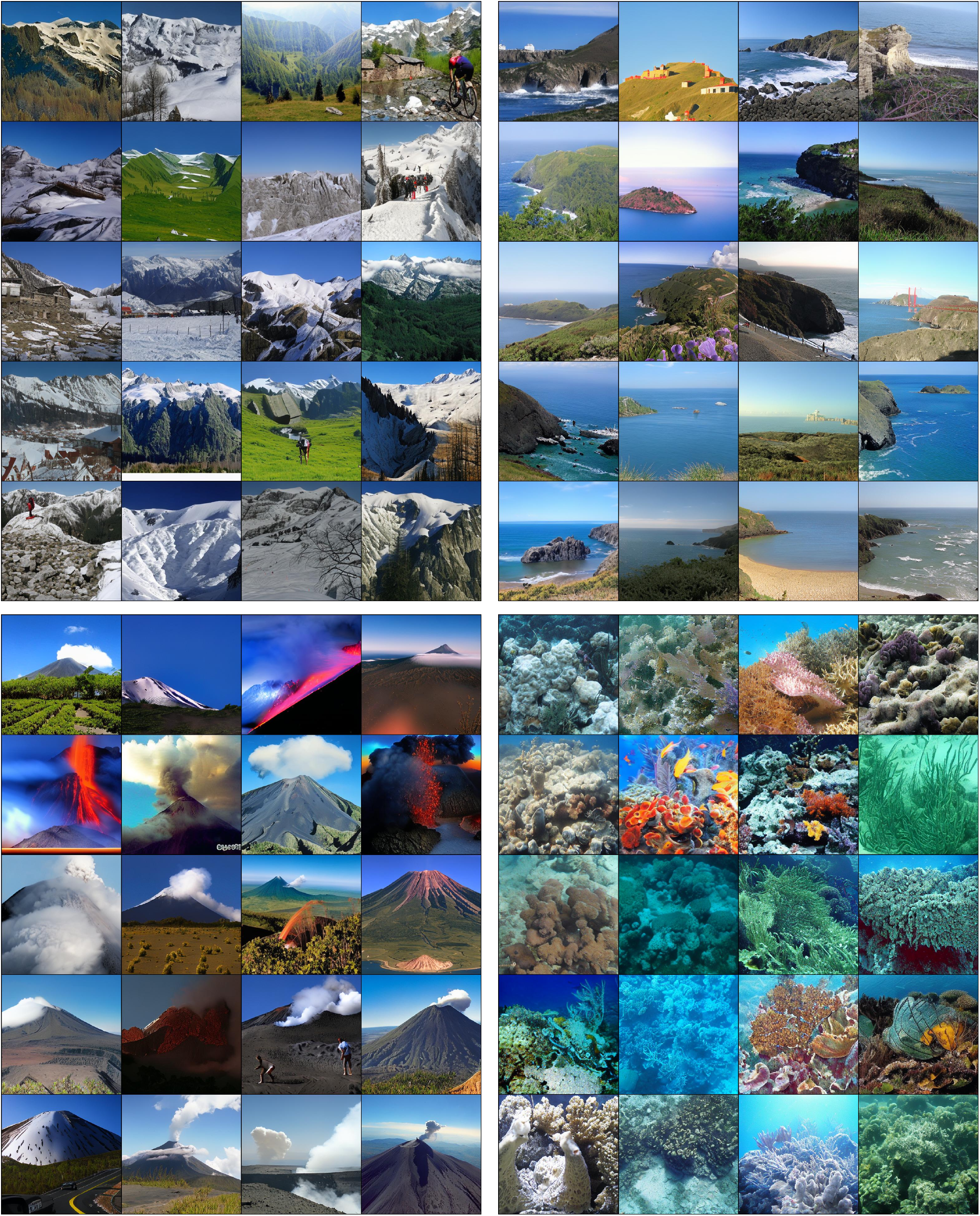}
\vspace{-0.1in}
\caption{The generated images for  alp, promontory, volcano, and coral reef by IAR-XXL with 4.0 CFG.}
\label{fig: generated_image4}
\vspace{-0.15in}
\end{figure*}

\hut{We exhibit more generated images from our model in Fig.~\ref{fig: generated_image4}$\sim$\ref{fig: generated_image1}, where the images are generated by The XXL-version with 4.0 CFG, with image size $384\times 384$. We show 16 classes of images, including alp, promontory, volcano, coral reef, sports car, balloon, convertible, space shuttle, castle, church, beacon, cinema, bridge, ocean liner, white stork, and Pomeranian. }

\section{\yr{Future Work}}
\label{sec:further improvement}
The main idea of our IAR is to ensure a high similarity between the predicted image embedding and the target embedding, so that even if the model incorrectly predicts the target token, the output image still closely resembles the target image. This can be naturally considered as a continuous constraint on the image embedding, aiming to minimize the distance between the predicted and target image embeddings. However, this approach cannot be easily applied to LLM-based image generation models due to the non-differentiable nature of the embedding quantization operation. Therefore, this paper relaxes the problem into a cluster-oriented token prediction problem, which can be easily integrated into the current autoregressive image generation model. We believe that in future work, employing this continuous constraint in autoregressive image generation may further enhance the model performance.


\clearpage

\begin{figure*}[t]
\centering
\includegraphics[width=1.0\textwidth]{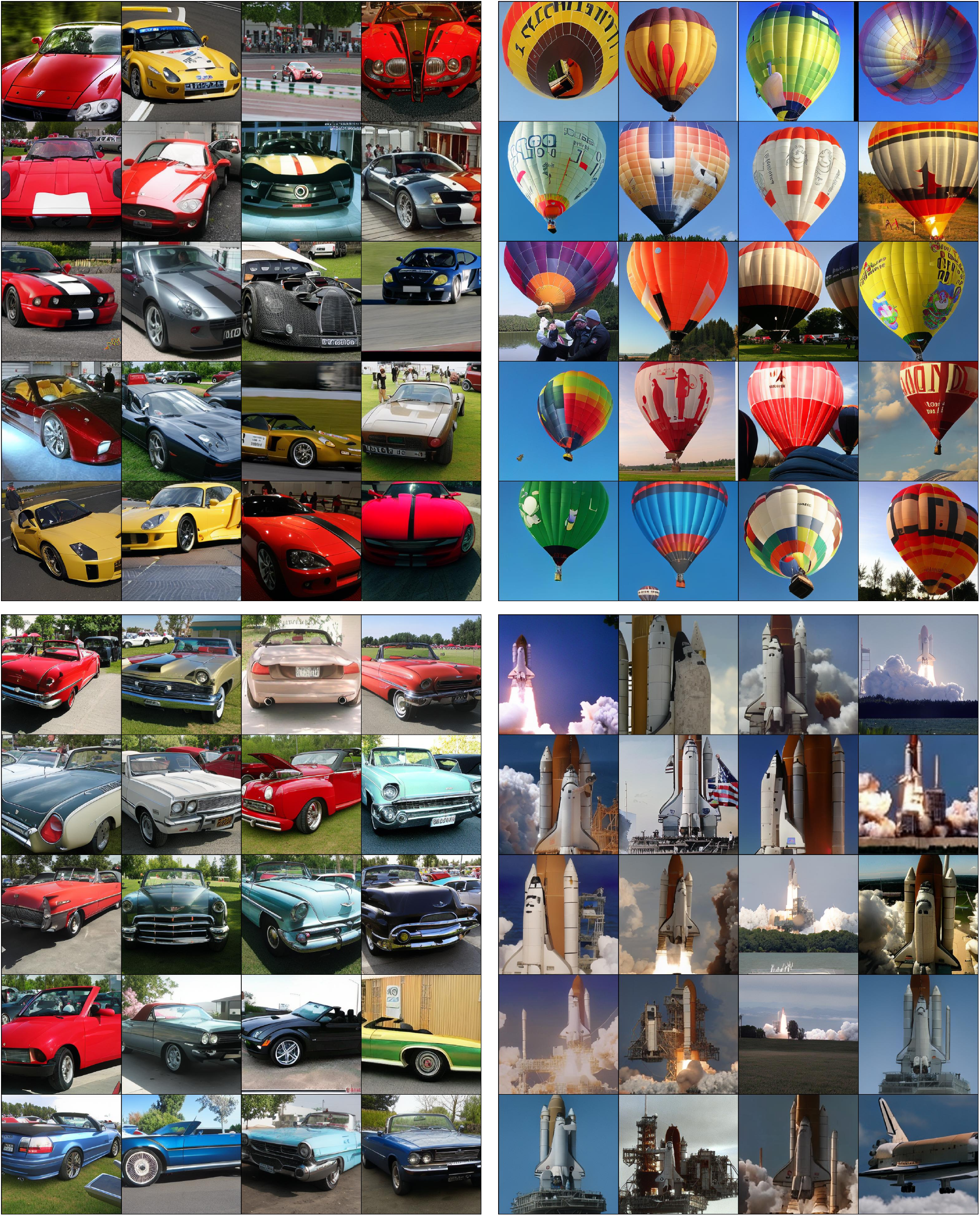}
\vspace{-0.1in}
\caption{The generated images for sports car, balloon, convertible, and space shuttle by IAR-XXL with 4.0 CFG.}
\label{fig: generated_image2}
\vspace{-0.15in}
\end{figure*}

\begin{figure*}[t]
\centering
\includegraphics[width=1.0\textwidth]{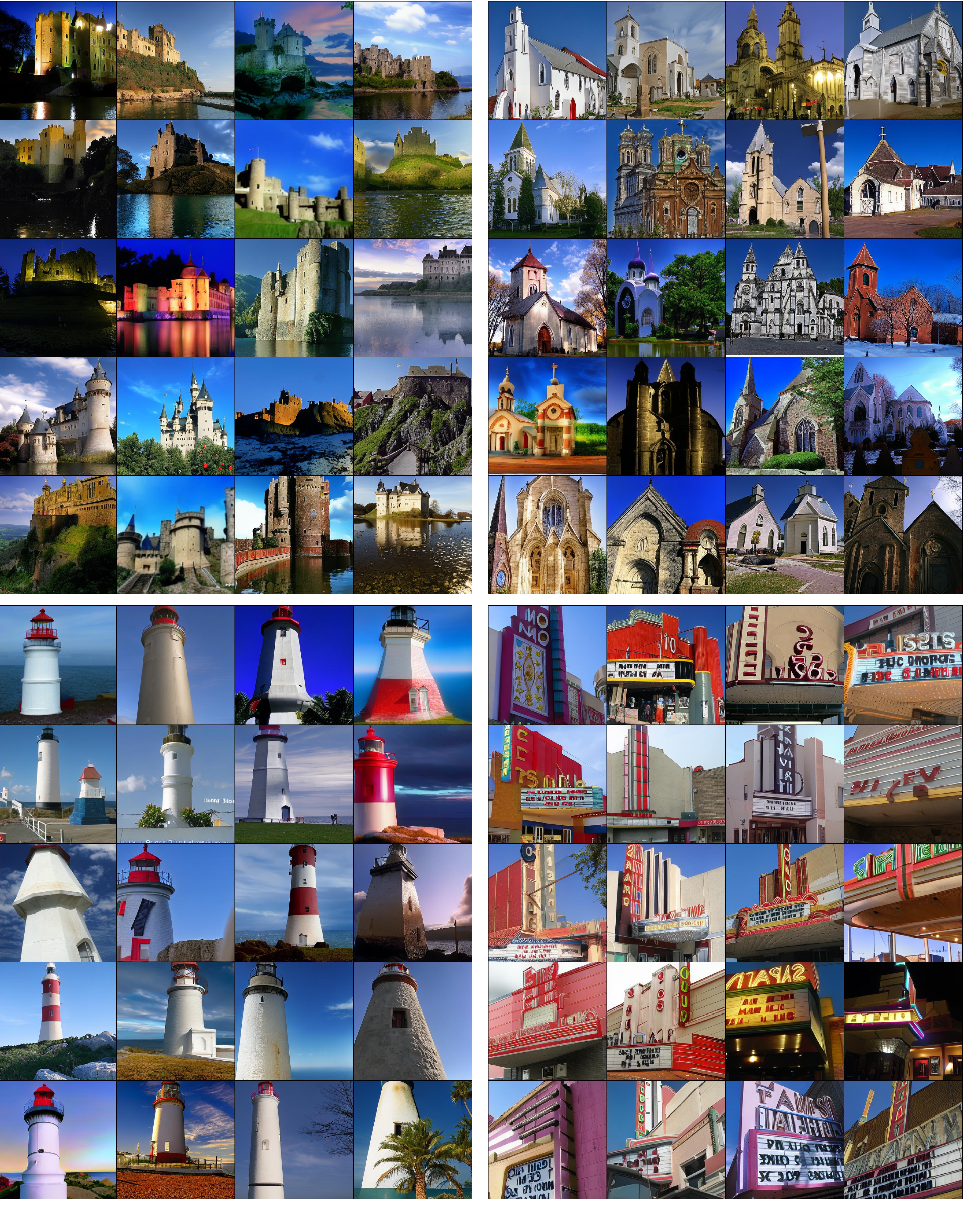}
\vspace{-0.1in}
\caption{The generated images for castle, church, beacon, and cinema by IAR-XXL with 4.0 CFG.}
\label{fig: generated_image3}
\vspace{-0.15in}
\end{figure*}

\begin{figure*}[t]
\centering
\includegraphics[width=1.0\textwidth]{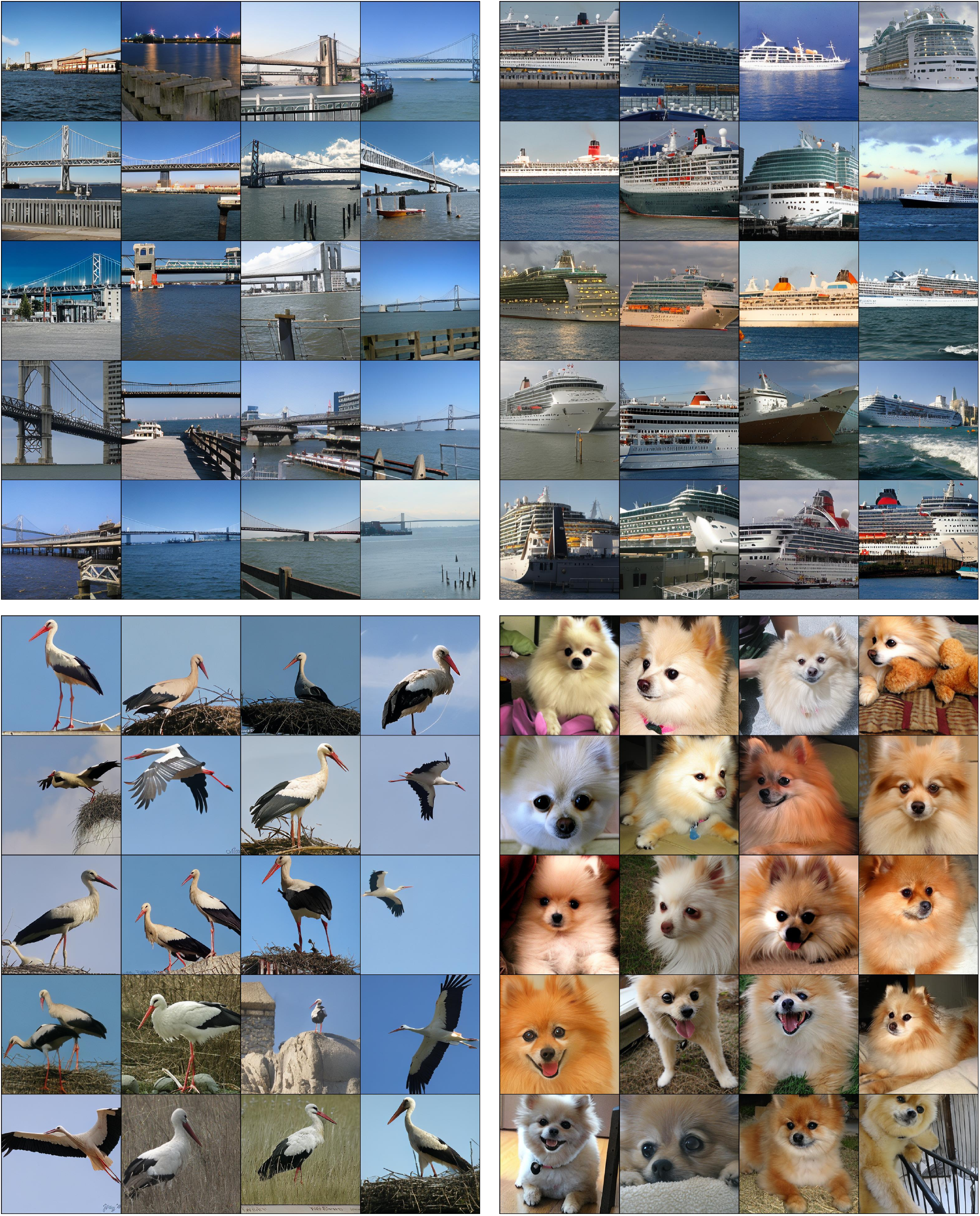}
\vspace{-0.1in}
\caption{The generated images for bridge, ocean liner, white stork, and Pomeranian by IAR-XXL with 4.0 CFG.}
\label{fig: generated_image1}
\vspace{-0.15in}
\end{figure*}

\end{document}

%% file: sec/1_intro.tex
\section{Introduction}
\label{sec:introduction}

\begin{figure}[t]
\centering
\includegraphics[width=0.48\textwidth]{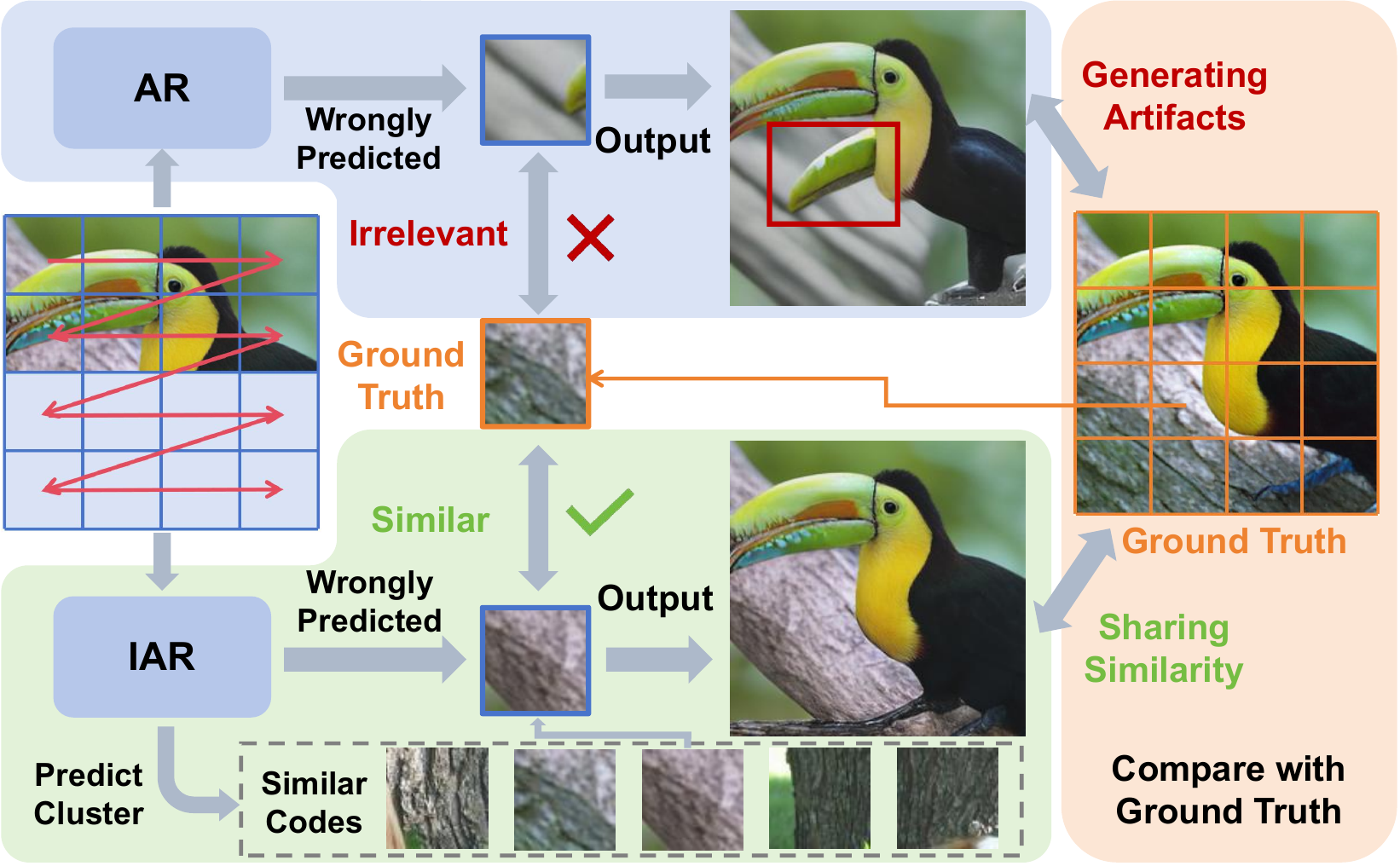}
\vspace{-0.15in}
\caption{
\zjn{When an autoregressive model predicts a wrong token, the previous methods~\cite{llamagen,VAR} may predict an irrelevant token that causes artifacts. Our IAR alleviates this issue by ensuring a high probability of the predicted token located in the correct cluster.}
}
\label{fig: motivation}
\vspace{-0.15in}
\end{figure}

With the development of generative models~\cite{ldm,goodfellow2020gan,ddpm}, a \yr{large} number of outstanding \yr{visual} generation \yr{methods} have emerged, achieving considerable success in image and video generation~\cite{ldm,stable_video_diffusion,animatediff}.
Recently, researchers have begun exploring the integration of images \yr{and text} to \yr{achieve a unified} multi-modal \yr{model for} image and text generation and understanding~\cite{transfusion,Chameleon}. 
Consequently, aligning image generation models with large language models (LLMs)~\cite{gpt} has become a \yr{key} research focus. Recent studies have made progress by quantizing images and employing autoregressive or mask-prediction methods to predict discrete image tokens~\cite{llamagen,VAR,maskgit,meissonic}, \yr{laying the} research foundation for \yr{a} unified world model \yr{in the} future.

Unlike diffusion~\cite{ddpm} or GAN~\cite{goodfellow2020gan} models that model image distribution in continuous space, autoregressive or 
\yr{masked image modeling (MIM)}
methods~\cite{llamagen,VAR,maskgit,meissonic} 
\yr{first convert images into discrete-valued tokens by image tokenizers,}
and then predict image tokens \yr{by autoregressive or MIM approach.} 
Specifically, autoregressive methods use the    ``next-token prediction" \yr{paradigm} of GPT~\cite{gpt}, while MIM 
methods adopt \yr{a similar} training strategy \yr{to the mask prediction in} BERT~\cite{bert}. 
\yr{These methods} draw inspiration from natural language models and transfer these techniques to image generation models, but rarely explore the fundamental differences between images and natural language. 

Due to the inherently discrete nature of words \yr{(each word comes from a finite vocabulary)}, 
different \yr{words can be directly mapped} to different indices through a lookup table, \yr{thereby generating} text by predicting the corresponding index \yr{of the} target text. \yr{The standard language modeling objective $p(x_t | x_1, \dots, x_{t-1})$ (each $x_t$ is an index) then can be interpreted as a classification task and accurately modeled by LLMs.
However, images are continuous-valued, and if \yrn{the same objective} is \yrn{applied to} image generation with $x_t$ as a real number, it is difficult to accurately model this probability density. Therefore, image tokenizers such as VQGAN~\cite{vqgan} are used to convert continuous images into discrete-valued tokens, and then the model predicts the image token index.}
\yr{Afterwards,} the corresponding image embedding is retrieved from the codebook and decoded into an image. 
\yr{A fundamental difference between image and text is that, in}
text generation, \yr{the predicted} text index can be \yr{directly mapped to} the corresponding word \yr{through} a lookup table\yr{;} whereas \yr{LLM-based} image generation \yr{essentially} requires the image embedding \yr{that} correspond\yr{s} to the index, \yr{and then decodes it in}to 
\yr{an} 
image. \yr{\emph{I.e.,} text only requires the index, while image requires the embedding corresponding to the index.}

Different from the token index \yr{which is} independent, \yrn{\emph{i.e.,}} the nearby \yr{indices have} no correlation \yrn{in word semantics}, the image embeddings are located in a continuous feature space, where similar embeddings may \yr{correspond to} similar image \yr{contents}. This paper first investigates the impact of image embedding correlations on \yr{the} generat\yr{ed} results, \yr{and finds} that \textit{similar embeddings convey similar information in the image space}\yr{:} When the embeddings of certain \yr{image} patches are \yr{replaced} by similar image embeddings, the \yr{decoded images} are nearly identical. This insight inspire\yr{s} us to leverage the similarity of image embeddings to improve \yr{the} existing LLM-based image generation techniques.


\hut{To capitalize on the correlation among image embeddings}, we propose \hut{\textbf{IAR}, an \textbf{I}mproved \textbf{A}uto\textbf{R}egressive Visual Generation Method, which can enhance} the training efficiency and generation quality of the LLM-based \yr{visual} generation model. 
\yr{Firstly}, we propose a \yr{\textbf{Codebook Rearrangement} strategy that uses} balanced K-means clustering algorithm to rearrange the \yr{embeddings in the} codebook into clusters \yr{of} equal sizes, where the embeddings in each cluster share high similarities and can \yr{be decoded into} images \yr{with} similar \yr{contents}. 
\yr{Secondly, we observe that if a token index is wrongly predicted as a similar embedding's index, the original token-oriented cross-entropy loss used in LLM will penalize this case, but the decoded image is actually not much different from the target image. To tolerate this case,}
based on the rearranged codebook, we further introduce a \textbf{Cluster-oriented Cross-entropy Loss}, which \yr{guides the model to} predict the \yr{correct} cluster that contains the target token, 
\yr{thereby providing} the model with a broader perspective and no longer confined to a single target token. 
Combined with the \yr{original} token-oriented cross-entropy loss, our model can consider both the target cluster and the target token. 
\yr{E}ven if the model predicts the wrong token index, with \yr{our} rearranged codebook and cluster-oriented loss, there is a high probability that the token \yr{is} located \yr{in} the target cluster, which ensure\yr{s} a high similarity between the output image and the target one, effectively improving the generation quality and robustness (Fig.~\ref{fig: motivation}).


\yr{We develop our model based on LlamaGen~\cite{llamagen}}
and compare with existing \yr{visual generation} methods based on GAN, Diffusion, Autoregressive (AR), and 
\yr{MIM models. Extensive} experiments demonstrate that, across different parameter scales of LlamaGen (ranging from 100M to 1.4B), our method consistently enhances the model's training efficiency and performance. 
For the same FID, our approach reduces LlamaGen's training time by half, and under the same number of training epochs, it \yrn{effectively} improves the model's generation quality.
Our method can be applied to \yr{various} LLM-based visual generation model\yr{s} with almost no additional training cost\yr{s,} and adheres to the scaling law\yr{,} provid\yr{ing} a promising direction for improving future LLM-based visual generation models.

Our main contributions can be summarized as four-fold:
\begin{itemize}
\item We 
propose \yrn{\textbf{IAR}, an Improved Autoregressive Visual Generation method}, which leverages the correlation of visual embeddings to enhance the training efficiency and performance of LLM-based visual generation models.
\item We introduce a \yr{\textbf{Codebook Rearrangement} strategy that uses} balanced K-means clustering algorithm to rearrange the image codebook, ensuring high similarity \yr{across} image embeddings within each cluster.
\item We propose a \textbf{Cluster-oriented Cross-entropy Loss} that relaxes the original token-oriented cross-entropy, ensuring that even if the model predicts the wrong token index, \yr{there is still a high probability that the token is in the correct cluster, thereby} 
generat\yr{ing} high-quality images.
\item Extensive experiments demonstrate the effectiveness of our method. Additionally, our approach can be applied to \yr{various} LLM-based visual generation model\yr{s}, adheres to the scaling law, and provides a robust improvement direction for future LLM-based visual generation models.
\end{itemize}

%% file: sec/2_related_work.tex
\section{Related Works}
\label{sec:related work}

\begin{figure*}[t]
\centering
\includegraphics[width=1.0\textwidth]{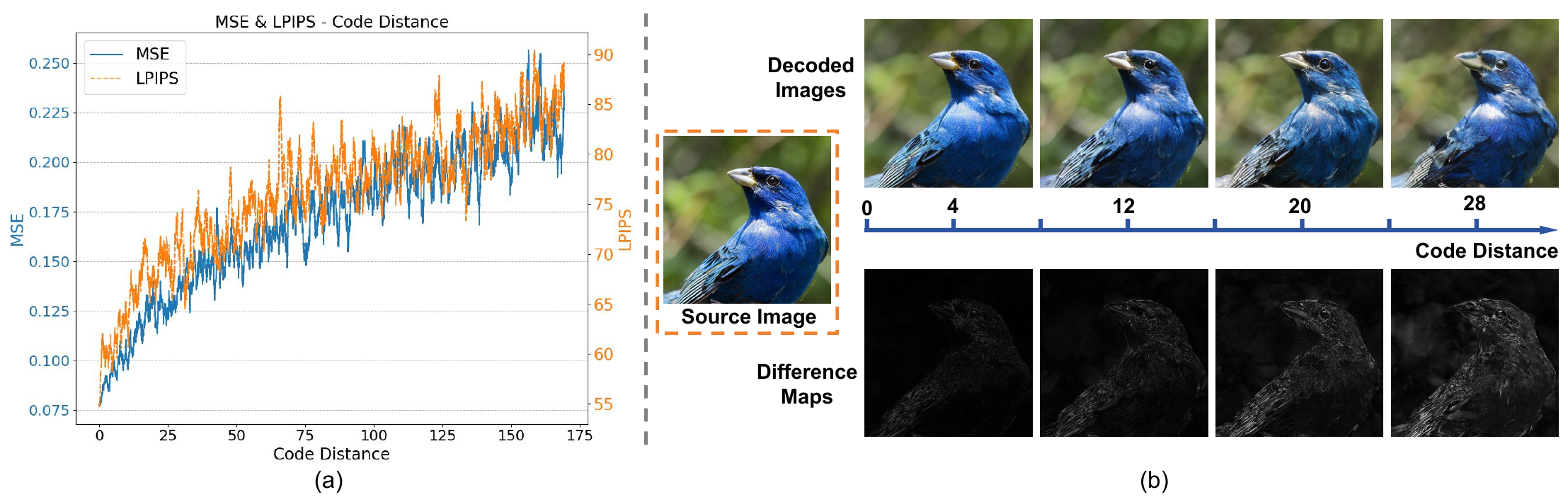}
\vspace{-0.2in}
\caption{(a) The MSE and LPIPS between the source image and the decoded image with different code distances. (b) The visualization of the images decoded from different code distances. 
\yrn{When the code distance is within a certain range (\emph{e.g.,} smaller than 12), the decoded image looks nearly identical to the source image.}
\yrn{We further}
make use of this property to improve the LLM-based \yrn{visual} generation model.}
\label{fig: code distance}
\vspace{-0.15in}
\end{figure*}

\subsection{Large language models}
Large language models (LLMs)~\cite{gpt,llama,bert} have achieved significant success in natural language processing (NLP).
The encoder-only structure~\cite{bert,electra,roberta}, exemplified by BERT~\cite{bert}, randomly masks portions of the text and predicts the masked tokens based on the unmasked ones. 
With the introduction of the GPT series~\cite{gpt,gpt1,gpt3,gpt4}, the decoder-only architecture gained popularity, demonstrating remarkable capabilities in language processing through next-token prediction. Additionally, open-source models such as LLaMA~\cite{llmsurvey} and PaLM~\cite{palm} have further advanced the NLP field. The recent success of LLMs in natural language processing has also inspired researchers to apply these models to visual generation, revealing significant potential in multi-modal generation tasks.


\subsection{\yrn{Visual} Generation Models}


{\bf Continuous-valued \yrn{Visual} Generation.}
Generative Adversarial Networks (GANs)~\cite{goodfellow2020gan,stylegan,vqgan,vit-vqgan,yi2024feditnet++}, consisting of a generator and discriminator, employ adversarial training to learn the target distribution. Diffusion Models~\cite{ddpm,ldm,stablediffusion3,adm,hu2023phasic,cdm,dit} iteratively add and remove noise to approximate the real data distribution, using a Markov process for the forward path and learning the reverse denoising process. These models have achieved significant success in image generation and have been extended to video generation~\cite{animatediff,stable_video_diffusion,xing2025dynamicrafter,hu2024motionmaster}, enhancing video quality. However, both GANs and Diffusion Models, which model continuous distributions, face challenges when integrating with discretely modeled LLMs. Thus, exploring discrete modeling methods for image generation is necessary.

{\bf Discrete-valued \yrn{Visual} Generation.} Discrete-space visual generation models are mainly divided into autoregressive~\cite{llamagen,vqgan,VAR} and masked image modeling (MIM)~\cite{maskgit,muse,meissonic} models, both requiring image quantization before generation. Autoregressive models follow the next-token generation approach of GPT~\cite{gpt,gpt4}, predicting subsequent tokens one by one based on previous image tokens. LLaMAGEN~\cite{llamagen}, a representative work, uses LLAMA~\cite{llama} to model the image generation process. VAR~\cite{VAR} further enhances generation capability by proposing a next-scale image generation process. MIM methods~\cite{llamagen,VAR,maskgit,meissonic}, inspired by BERT~\cite{bert}, randomly mask parts of the image tokens and predict the masked tokens using the unmasked ones. These methods directly draw inspiration from LLMs and rarely explore the fundamental differences between images and language, which may lead to suboptimal utilization of vision generation
 capabilities within the LLM framework. In contrast, our model makes full use of the image property, which effectively improves the training efficiency and model performance in LLM-based image generation.

%% file: sec/3_method.tex
\section{Method}
\label{Method}






\label{sec:image token similarity}

\begin{figure*}[t]
\centering
\includegraphics[width=1.0\textwidth]{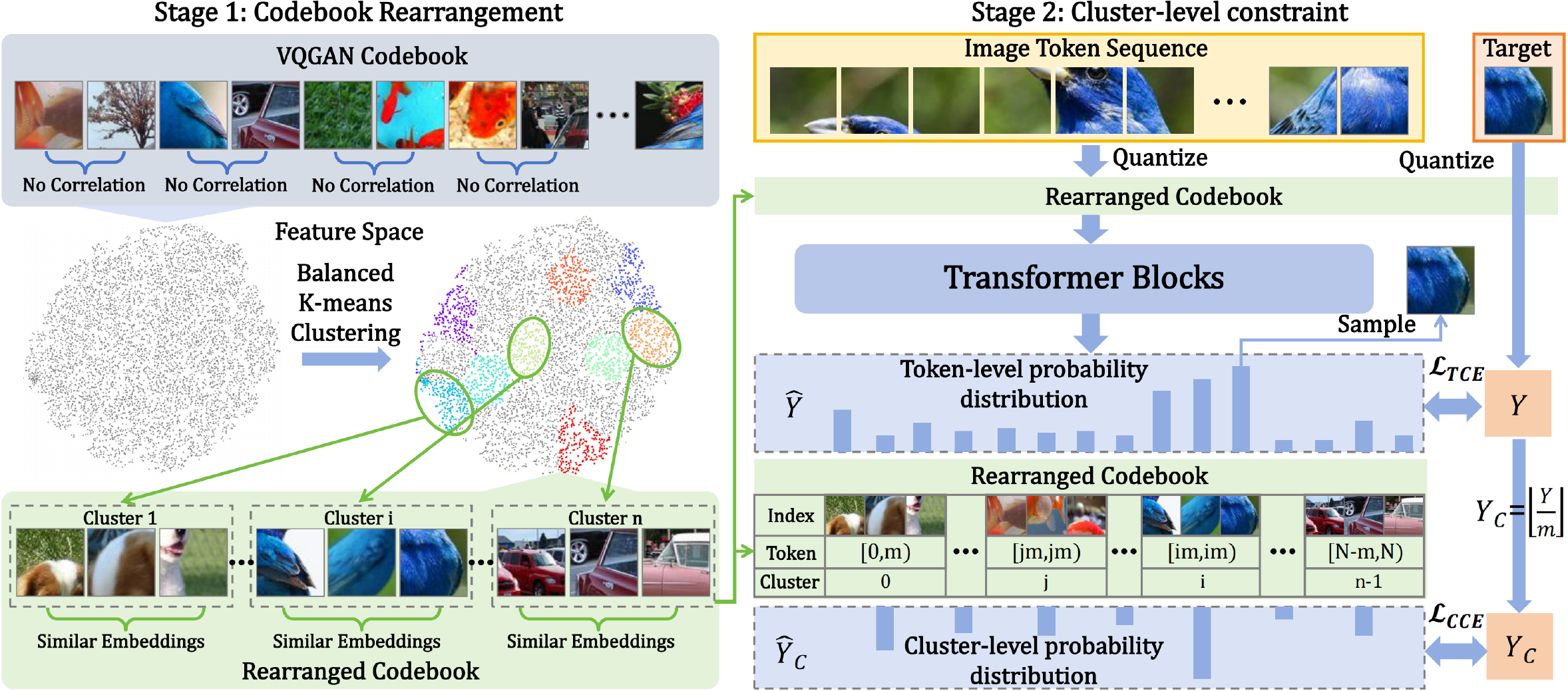}
\vspace{-0.15in}
\caption{Model framework: 1) \yrn{\textbf{Codebook Rearrangement}: we} first \yrn{use a} balanced K-means clustering method \yrn{to} rearrange the codebook, which \yrn{divides} the codebook into $n$ clusters\yrn{, with} the image codes in each cluster shar\yrn{ing} a high similarity. 
2) \yrn{\textbf{Cluster-oriented Constraint}:} During 
\yrn{the training process,}
we first quantize the image patches \yrn{using} the rearranged codebook. For the output probability distribution $\hat{Y}$, we further compute the \yrn{cluster-level} distribution $\hat{Y}_C$ by applying \yrn{LogSumExp} operation for the probabilities in each cluster $\hat{Y}_{\yrn{jm}}\sim \hat{Y}_{\yrn{(j+1)m-1}}$. 
\yrn{Then we compute} the cluster-oriented cross-entropy loss $\mathcal{L}_{CCE}$ \yrn{apart from} the token-oriented cross-entropy loss $\mathcal{L}_{TCE}$\yrn{, which ensures a high probability of the predicted token located in the correct cluster, thereby enhancing generation quality.}
}
\label{fig: Model framework}
\vspace{-0.1in}
\end{figure*}

\hut{
\yr{Observing that an inherent difference between LLM-based image generation and text generation lies in that text generation only requires a predicted text index that can be directly mapped into the corresponding word, while image generation essentially requires the image embedding that corresponds to the index, and then decodes it into an image, we further explore the characteristics of the image embedding space.}
We conduct \yr{a} detailed analysis on the similarity of the image embeddings \yr{in the codebook learned by image tokenizers (Sec.~\ref{sec:image token similarity}), and} find that \yr{similar embeddings convey similar information in the image space:} 
When the embeddings of certain \yr{image} patches are \yr{replaced} by similar image embeddings, the \yr{decoded images} are nearly identical.} 
Therefore, we leverage this property to enhance the LLM-based image generation model. 

We first propose a \yr{\textbf{Codebook Rearrangement} strategy that uses a} balanced K-means clustering method to rearrange \yr{the embeddings in} the codebook into clusters of equal sizes (Fig.~\ref{fig: Model framework} (left)\yr{, Sec.~\ref{sec:code_rearrangement}), where} 
the image embeddings in each cluster share a high similarity. 
\hut{Once the model predict\yr{s} the \yr{correct} cluster index \yr{(even if the token index is wrong), the corresponding embedding is located in the correct cluster and is thus similar to the target one, then} the output images \yr{will} be quite similar to the \yr{target image}. 
To enable the model with the ability to predict the \yr{correct} cluster index, we further propose a \textbf{Cluster-oriented Cross-entropy Loss} \yr{(Fig.~\ref{fig: Model framework} (right), Sec.~\ref{sec:cluster-oriented-loss}).}}
\yr{Our} rearranged codebook and cluster-oriented cross-entropy loss \yr{ensure that}, 
even if the model predicts \yr{incorrect tokens, 
there is a high probability of correctly predicting} the cluster \yr{indices,}
\yr{which} ensures high-quality generation result\yr{s}, largely improv\yr{ing} the training efficiency and robustness of the LLM-based image generation model.

\subsection{Analysis on Image \yr{Embedding} Similarity}

We first analyze the similarity of image embeddings in the codebook \yr{learned by image tokenizers,} and find that similar embeddings (those that are close in distance) represent similar image information.
\yr{\emph{I.e.,}} when the embeddings \yr{of} each \yr{image} patch are replaced by other similar embeddings, the content of the \yr{decoded} image remains almost unchanged. 
To verify this \yr{image embedding similarity property,} 
we conduct detailed experiments on the codebook of VQGAN~\cite{vqgan}.

Specifically, an image $x$ \yr{is quantized into discrete tokens by: 1)} us\yr{ing an} image encoder $E$ \yr{(\emph{e.g.,}} from VQGAN~\cite{vqgan}) to extract features, obtaining $h \times w$ image features of dimension $C$, denoted as $\hat{z}=E(x) \in \mathcal{R}^{h \times w \times C}$\yr{; 2)} quantiz\yr{ing each feature $\hat{z}^{(i,j)}$ into the code index $q^{(i,j)}$ of the nearest code in} the codebook $\mathcal{Z} \yr{\in \mathcal{R}^{N \times C}}$: 
\begin{equation}
    \yr{q^{(i,j)}}=\arg\min\limits_{z_k\in \mathcal{Z}}\|\hat{z}^{(i,j)}-z_k\| \yr{\in [0,N)},
\end{equation}
\yr{where $z_k$ is the $k$-th embedding in codebook $\mathcal{Z}$, and $\| \cdot \|$ measures the Euclidean distance. And the quantized feature $z_q \in \mathcal{R}^{h \times w \times C}$ is obtained by looking up each $q^{(i,j)}$ in $\mathcal{Z}$.}

To measure the distance between 
\yr{two embeddings $z_i$, $z_j$}
in the codebook, we define the ``code distance"
\yr{$D(z_i, z_j)$}:
\begin{equation}
\begin{aligned}
    \yr{D(z_i, z_j):=d, \quad z_j = [ \text{sort}(\{\| z_k-z_i \|\}) ]_d,}
\end{aligned}
\end{equation}
\yr{calculated by measuring the Euclidean distance of each $z_k$ to $z_i$, sorting the distances, and $z_j$ is the $D(z_i, z_j)$-th closest embedding to $z_i$ among all embeddings in the codebook. A lower ``code distance" indicates the two embeddings are closer in the embedding space.}

\yr{Given an input image $x$, the quantized embedding $z_q$, and a generated image $\hat{x}'=G(z_q')$ decoded from $z_q'$, we evaluate the similarity of the two images}
at different code distances $D(z_q,z_q')$. 
In Fig.~\ref{fig: code distance} (a), we measure the image similarity \yrn{by} MSE distance and LPIPS~\cite{lpips}, where lower \yrn{ scores indicate} \yr{higher} image similarity. 
\yrn{Results show: 1)} as the code distance increases, 
\yr{the image distance also increases;} 
\yrn{2)} but \yr{when the code distance is within a certain range \yrn{(\emph{e.g.,} smaller than $12$)},} the image distance \yrn{(MSE $0.104$ and LPIPS $66.44$)} does not increase too much compared to the reconstructed image \yr{$G(z_q)$} (whose code distance is $0$ and has a $0.076$ MSE and $55.8$ LPIPS). 
We \yrn{further} visualize the images decoded from different code distances in Fig.~\ref{fig: code distance}(b)\yrn{, which shows} 
\yr{the decoded images \yrn{with lower code distances (\emph{e.g.,} smaller than $12$)}} 
\yr{look \yrn{nearly identical} to the source image}
and have a good \yr{visual} quality. 
This indicates 
that even if the \yr{predicted \yrn{token} index} 
is not \yr{the accurate target index,} 
\yr{as long as} the code distance \yr{between the corresponding embedding} is \yr{within a certain range}, 
the \yr{decoded} image \yr{is} similar to the desired one \yr{and shows a good visual quality.}
Therefore, we make use of this property to improve the generation quality and stability of the LLM-based \yr{visual} generation model.

\begin{figure}[t]
\centering
\resizebox{0.45\textwidth}{!}{
\begin{minipage}{0.5\textwidth}
\begin{algorithm}[H]
\caption{Balanced k-means Clustering}
\renewcommand{\algorithmicrequire}{\textbf{Input:}}
\renewcommand{\algorithmicensure}{\textbf{Output:}}
\begin{algorithmic}[1]
\REQUIRE Codebook $\mathcal{Z}$, number of clusters $n$
\ENSURE The rearranged codebook \yr{$\hat{\mathcal{Z}}$} with local similarity

\STATE \textbf{Initialize:}
\STATE \quad Compute the cluster size $m \gets |\mathcal{Z}| / n$
\STATE \quad Randomly select $n$ codes as initial centroids $\{c_j\}_{j=1}^n$
\STATE \quad Initialize iteration counter $iter \leftarrow 0$

\REPEAT
    \STATE Compute the minimum distance $d_i$ from \yr{$z_i$} to $\{c_j\}_{j=1}^n$

    \STATE Sort \yr{$\{z_i\}$} in ascending order of $d_i$ to form $\mathcal{Z}'$

    \FOR{each data point $\yr{z_i} \in \mathcal{Z}'$}
        \STATE Compute the distance from \yr{$z_i$} to each centroid $c_j$
        \STATE Assign \yr{$z_i$} to the nearest cluster whose size does not exceed $m$
    \ENDFOR

    \FOR{each \yr{cluster} centroid $c_j$}
        \STATE Update $c_j$ \yr{to} 
        the \yr{average} of all \yr{embeddings} assigned to \yr{cluster $j$} 
    \ENDFOR

    \STATE Increment iteration counter $iter \leftarrow iter + 1$
\UNTIL{convergence or reaching $max\_iters$}

\FOR{each \yr{cluster} centroid $c_j$}

\STATE 
\yr{Assign the embeddings in cluster $j$ to $\hat{\mathcal{Z}}_{[jm, (j+1)m)}$}
\ENDFOR
\RETURN The rearranged codebook $\hat{\mathcal{Z}}$

\end{algorithmic}
\label{alg:k-means}
\end{algorithm}
\end{minipage}}
\end{figure}

\subsection{Codebook Rearrangement}
\label{sec:code_rearrangement}

{\bf Rearrangement Target.} In a codebook \yr{learned} from \yr{the} 
VQGAN~\cite{vqgan} training \yr{process}, \yr{the embeddings} are often randomly distributed, 
with adjacent \yr{embeddings} having no particular correlation\yr{, \emph{i.e.,} Euclidean distances between adjacent embeddings can be either close or far}. 
Therefore, we first propose a \yr{Codebook Rearrangement strategy} to reorder the codebook, so that the \yr{similarity} between adjacent \yr{embeddings} is maximized, ensuring that the surrounding \yr{embeddings} of any given \yr{embedding} exhibit \yr{high} similarity.

Specifically, denote the codebook as $\mathcal{Z}=\{z_i\}_{i=1}^N$, which contains $N$ quantized \yr{embeddings} $z_i$. To maximize the similarity between the adjacent \yr{embeddings}, we aim to find a surjective \yr{mapping} $M(\cdot)$ that satisfies:
\begin{equation}
    \begin{aligned}
        M=\arg\min\limits_{M} \sum_{i=1}^{N-1} \yr{\|}z_{M(i)},z_{M(i+1)}\yr{\|}.
    \end{aligned}
\end{equation}

\yr{\emph{I.e.,} after reordering each embedding $z_i$ to index $M(i)$, the sum of distances between adjacent embeddings is minimized.}
And $\hat{\mathcal{Z}}=M(\mathcal{Z})$ is the rearranged codebook.

However, this optimization can be reduced to the Hamiltonian path problem (see \#Suppl), and solving such a problem is NP-hard. Therefore, we try to relax this problem into a solvable form.

{\bf Constraint Relaxation.} We relax this optimization \yr{problem} into a much easier one, where we only need to ensure the embeddings in a range \yr{are similar.} 
Therefore, we split the codebook into $n$ clusters, with each cluster containing $m=\frac{N}{n}$ 
embeddings that are similar to each other. 
\yr{Then we reorder the codebook so that the embeddings in the same cluster have close indices, with the indices of}
cluster $j$ 
\yr{are within the range $[jm, (j+1)m$\hut{)}.}

{\bf Balanced K-means Clustering.}  
\yr{To cluster the embeddings in the codebook,}
we \yrn{design} a balanced K-means clustering method to uniformly divide the codebook into $n$ clusters, \yrn{with} the \yr{embeddings} in each cluster close to each other.

As shown in Alg.~\ref{alg:k-means}, we first randomly select $n$ codes as the initial centroids $\{c_j\}_{j=1}^n$. 
Then, we \yr{iteratively} update each cluster. 
In each iteration, \yrn{1)} we first compute the minimum distance $d_i$ from each \yr{embedding $z_i$} to the $n$ centroids $\{c_j\}_{j=1}^n$. 
\yrn{2)} We sort \yr{$\{z_i\}$} in ascending order of $d_i$ to form $\mathcal{Z}'$ 
(\emph{i.e.}, \yr{embeddings} closer to the nearest centroid \yr{has \yrn{smaller} index}). \yrn{This sorting and reordering step ensures that the embeddings closer to the nearest centroid are allocated first, and the farther ones are allocated later, which is necessary in cluster size balanced clustering.} 
\yrn{3)} Then, 
\yr{for each $z_i$ in} the ordered codebook $\mathcal{Z}'$, we compute \yr{its} distance to each centroid $c_j$. 
\yrn{4)} We assign \yr{$z_i$} to the nearest cluster whose size does not exceed $m$ (where $m=\frac{N}{n}$). 
\yrn{5)} Finally, we update \yr{each cluster} centroid \yr{$c_j$ to the average of the embeddings assigned to that} 
cluster, and repeat the \yr{above} process until it converges or reaches the pre-defined maximum iteration. 
\yr{After obtaining} the final cluster \yr{centroids} $c_j$ \yr{and $n$ clusters,} where each cluster contains $m$ \yr{embeddings}, we rearrange the codebook $\hat{\mathcal{Z}}$ by assigning the 
\yr{embeddings in cluster $j$}
to indices \yr{$[jm, (j+1)m)$}. 


\subsection{Cluster-oriented \yr{Visual} Generation}
\label{sec:cluster-oriented-loss}

{\bf \yr{Analysis on} Token-oriented Cross-entropy Loss.} In LLM-based \yr{visual} generation models~\cite{llamagen,VAR}, cross-entropy loss is the most commonly used \yr{loss in the} training process. Denote the ground truth one-hot vector as $Y$ (\yrn{$Y$ is a $N$-dimensional vector, with} $Y_{i=y}=1$ and $Y_{i\neq y}=0$, where $y$ is the class label \yrn{of the token}), and the predicted \yrn{probability distribution} as $\hat{Y} \yrn{\in \mathcal{R}^N (\sum \hat{Y}_{i}=1)}$. The token-oriented cross-entropy loss $\mathcal{L}_{TCE}$ \yr{is formulated} as:
\begin{equation}
    \begin{aligned}
        \mathcal{L}_{TCE}=-\sum_{i=1}^n Y_i\log\hat{Y}_i.
    \end{aligned}
\end{equation}
\yr{However, we observe that if a token index is wrongly predicted as a similar embedding's index, this token-oriented cross-entropy loss will penalize the case, but the decoded image is actually not much different from the target image.}

\yr{To tolerate this case, we leverage} 
the rearranged codebook, \yr{where} the \yr{embeddings} in the same cluster 
\yr{are close to each other.}
In Sec\yr{.~\ref{sec:image token similarity}}, it has been demonstrated that even if some image token \yr{indices} are predicted incorrectly, as long as \yr{the embeddings of the} predicted tokens are not far from the \yr{embeddings of the} target tokens, the semantic integrity and quality of the generated image will not be significantly affected. 
Therefore, leveraging the reordered codebook from Sec\yr{.~\ref{sec:code_rearrangement}}, a natural idea is to \yr{guide} the model \yr{to} first predict the cluster and then locate the specific token within that cluster. Since the number of clusters ($n$) is much smaller than the size of the codebook ($N$), predicting the cluster \yr{of a token} 
is a much easier task than directly predicting a specific token. 
Once the cluster is accurately predicted, even if the specific token prediction is incorrect, the quality of the generated image can still be maintained.

{\bf Cluster-oriented Cross-entropy Loss.} To guide the model \yr{to predict the correct} cluster of the next token, 
we propose a cluster-wise cross-entropy loss $\mathcal{L}_{C}$. Specifically, we first derive the ground truth cluster label $y_c$ from the class label $y$\footnote{\yr{After codebook rearrangement, the class label has been updated accordingly.}} by $y_c=\lfloor \frac{y}{m} \rfloor$, where $m$ is the number of tokens in \yr{each} cluster. 
Then, to form a new \yrn{cluster-level} probability distribution $\hat{Y}_{C} \yrn{\in \mathcal{R}^n (\sum \hat{Y}_{C,i}=1)}$ from the \yrn{token-level} probability distribution $\hat{Y}$, we employ \hut{the following} operation to \yr{calculate} the probability of each cluster\hut{, where \yrn{within a cluster,} the sample\yrn{s} with larger probability contribute more to the cluster probability}:
\begin{equation}
\begin{aligned}
    \hat{Y}_{C,j}&= \nicefrac{\sum\limits_{i=jm}^{(j+1)m-1} exp(\hat{Y}_{i})} {\sum\limits_{i=1}^{\yrn{N}} 
    exp(\hat{Y}_i)},\yrn{\quad j=1,\ldots,n}
\end{aligned}
\end{equation}
\yrn{where $\hat{Y}_{C,j}$ represents the probability that the generated token belongs to cluster $j$.}

With the ground-truth \yrn{cluster label} \yr{$y_c$} and the predicted \yrn{cluster-level} probability distribution $\hat{Y}_{C,i}$, our cluster-oriented cross-entropy loss $\mathcal{L}_{CCE}$ \yr{is} formulated \yr{as}:
\begin{equation}
\begin{aligned}
    \mathcal{L}_{CCE}=-\sum_{\yrn{j}=1}^n Y_{C,\yrn{j}}\log\hat{Y}_{C,\yrn{j}},
\end{aligned}
\end{equation}
where $Y_C$ is the one-hot vector spanned by the label $y_c$.
\yr{With our cluster-oriented loss, even if the model predicts the wrong token index, there is a high probability that the token is located in the target cluster, which ensures a high similarity between the output image and the target one, effectively improving the generation quality and robustness.}

{\bf Final Loss Function.} The final loss function $\mathcal{L}$ combines both the token-oriented cross-entropy loss $\mathcal{L}_{TCE}$ and the cluster-oriented cross-entropy loss $\mathcal{L}_{CCE}$:
\begin{equation}
\mathcal{L}=\mathcal{L}_{TCE}+\lambda\mathcal{L}_{CCE}.
\end{equation}


%% file: sec/4_experiment.tex
\section{Experiments}
\label{experiment}

\begin{table}[t]
\centering
\resizebox{0.48\textwidth}{!}{
\begin{tabular}{c|lc|cccc}
\toprule
Type & Model & \#Para. & FID$\downarrow$ & IS$\uparrow$ & Precision$\uparrow$ & Recall$\uparrow$  \\
\midrule
\multirow{3}{*}{GAN}   & BigGAN~\citep{biggan}  & 112M   & 6.95  & 224.5       & 0.89 & 0.38 \\
 & GigaGAN~\citep{gigagan}  & 569M    & 3.45  & 225.5       & 0.84 & 0.61  \\
 & StyleGan-XL~\citep{stylegan-xl} & 166M    & 2.30  & 265.1       & 0.78 & 0.53   \\
\midrule
\multirow{4}{*}{Diffusion} & ADM~\citep{adm}  & 554M       & 10.94 & 101.0        & 0.69 & 0.63    \\
 & CDM~\citep{cdm}   & $-$       & 4.88  & 158.7       & $-$  & $-$   \\
 & LDM-4~\citep{ldm} & 400M     & 3.60  & 247.7       & $-$  & $-$  \\
 & DiT-XL/2~\citep{dit}  & 675M  & 2.27  & 278.2       & 0.83 & 0.57   \\
\midrule
\multirow{2}{*}{Mask.} & MaskGIT~\citep{maskgit}  & 227M   & 6.18  & 182.1        & 0.80 & 0.51  \\
 & MaskGIT-re~\citep{maskgit} & 227M\    & 4.02  & 355.6        & $-$ & $-$ \\
\midrule
\multirow{7}{*}{AR} & VQGAN~\citep{vqgan} & 227M & 18.65 & 80.4         & 0.78 & 0.26    \\
 & VQGAN~\citep{vqgan}    & 1.4B   & 15.78 & 74.3   & $-$  & $-$     \\
 & VQGAN-re~\citep{vqgan}  & 1.4B  & 5.20  & 280.3  & $-$  & $-$     \\
 & ViT-VQGAN~\citep{vit-vqgan} & 1.7B & 4.17  & 175.1  & $-$  & $-$        \\
 & ViT-VQGAN-re~\citep{vit-vqgan}& 1.7B  & 3.04  & 227.4  & $-$  & $-$     \\
 & RQTran.~\citep{rq}       & 3.8B  & 7.55  & 134.0  & $-$  & $-$     \\
 & RQTran.-re~\citep{rq}    & 3.8B & 3.80  & 323.7  & $-$  & $-$    \\
\midrule
\multirow{4}{*}{AR} & LlamaGen-B~\cite{llamagen}& 111M & 5.46 & 193.6 & 0.83 & 0.45\\
 & LlamaGen-L~\cite{llamagen}   & 343M & 3.29 & 227.8 & 0.82 & 0.53 \\
 & LlamaGen-XL~\cite{llamagen}  & 775M & 2.63 & 244.1 & 0.81 & 0.58 \\
 & LlamaGen-XXL~\cite{llamagen}  & 1.4B & 2.34 & 253.9 & 0.80 & 0.59 \\
\midrule
\multirow{4}{*}{Ours} & IAR-B& 111M & 5.14 & 202.0 & 0.85 & 0.45\\
 & IAR-L  & 343M & 3.18 & 234.8 & 0.82 & 0.53 \\
 & IAR-XL  & 775M & 2.52 & 248.1 & 0.82 & 0.58 \\
& IAR-XXL (CFG=1.7)  & 1.4B & \textbf{2.19} & 278.9 & 0.81 & 0.58 \\
& IAR-XXL (CFG=2.5)  & 1.4B & 3.74 & \textbf{362.0} & 0.86 & 0.54 \\
\bottomrule
\end{tabular}}
\vspace{-0.1in}
\caption{\textbf{Comparison between different types of image generation model on class-conditional ImageNet 256$\times$256 benchmark} with FID, IS, precision, and recall.
}
\vspace{-0.15in}
\label{tab:main}
\end{table}

\subsection{Experiment\yrn{al} settings}
In this paper, we choose LLamaGen~\cite{llamagen} as the main baseline and apply our method \yrn{to} it. We follow the official training details and keep the hyperparameters the same as the official ones. We conduct experiments on ImageNet~\cite{deng2009imagenet}. To evaluate the model performance, we generate 50K images with randomly selected labels. Then, we compute the FID~\cite{fid}, IS~\cite{inception_score}, precision, and recall~\cite{precision_and_recall} for the generated data (see \#Suppl for details).

\begin{table*}[t]
\centering
\resizebox{0.8\textwidth}{!}{
\begin{tabular}{c|lc|cccc|cccc}
\toprule
\multirow{2}{*}{Tokens} & \multicolumn{1}{c}{\multirow{2}{*}{Model}} & \multicolumn{1}{c}{\multirow{2}{*}{\#Para.}} & \multicolumn{4}{c}{50 epoch}                                                                                 & \multicolumn{4}{c}{300 epoch}                                                                                \\
                              & \multicolumn{1}{c}{}                       & \multicolumn{1}{c}{}                      & \multicolumn{1}{c}{FID$\downarrow$} & \multicolumn{1}{c}{IS$\uparrow$} & \multicolumn{1}{c}{Precision$\uparrow$} & \multicolumn{1}{c}{Recal$\uparrow$} & \multicolumn{1}{c}{FID$\downarrow$} & \multicolumn{1}{c}{IS$\uparrow$} & \multicolumn{1}{c}{Precision$\uparrow$} & \multicolumn{1}{c}{Recal$\uparrow$} \\
                              \midrule
\multirow{10}{*}{16 $\times$ 16}       & LlamaGen-B                                 & 111M                                      & 7.22                    & 178.3                & 0.86                         & 0.38                     & 5.46                   & 193.6                & 0.84                         & \textbf{0.46}                     \\
                              & LlamaGen-L                                 & 343M   & 4.20                   & 200.0                & 0.82                         & 0.51                                                       & 3.80                   & 248.3                 & 0.83                         & \textbf{0.52}                      \\
                              & LlamaGen-XL                                & 775M                                      &         3.39                  &         227.1             &          0.81                   &    0.54                          &             -            &            -            &       -                        &               -            \\
                              & LlamaGen-XXL                               & 1.4B                                      &        3.09                 &    253.6                    &        0.83                       &     0.53                      &                -         &           -             &    -                           &               -            \\
                              \cline{2-11}
                              & IAR-B                                & 111M                                      &\textbf{6.90}                    & \textbf{179.2}                 & \textbf{0.86}                        & \textbf{0.40}                   & \textbf{5.14}                   & \textbf{202.0}                & \textbf{0.85}                       & 0.45                    \\
                              & IAR-L                                & 343M                                      & \textbf{4.10}
                              
                              & \textbf{207.1}                &\textbf{0.82}                        & \textbf{0.51}                    & \textbf{3.40}                  & \textbf{271.3}                 & \textbf{0.84}                        & 0.51                     \\
                              & IAR-XL                               & 775M                                      & \textbf{3.36}                    & \textbf{228.9}               & \textbf{0.82}                        &\textbf{0.54}                     &                 -        &          -              &        -                       &             -              \\
                              & IAR-XXL                              & 1.4B                                      &        \textbf{3.01}                 &   \textbf{257.4}                     &        \textbf{0.83}                       &        \textbf{0.53}                   &               -          &          -              &      -                         &                -           \\
                              \midrule
\multirow{8}{*}{24 $\times$ 24}        & LlamaGen-B                                 & 111M                                      & 8.31                   &\textbf{154.7}               & 0.84                         & 0.38                     & 6.09                   & 182.5                & 0.84                         & 0.42                     \\
                              & LlamaGen-L                                & 343M                                      &             4.61			
            &      191.4                  &             \textbf{0.82}                  &  0.50                         & 3.29                   & 227.8                 & 0.82                        & 0.53                    \\
                              & LlamaGen-XL                                & 775M                                      & 3.24                   & \textbf{245.7}                 & \textbf{0.83}                         & 0.53                     & 2.63                   & 244.1                & 0.81                         & 0.58                     \\
                              & LlamaGen-XXL                               & 1.4B                                      &      2.89                                            &         236.2                      &   0.80                        &        0.56                  &     2.34    &    253.9           &       0.81                        &     \textbf{0.60}                      \\ \cline{2-11}
                              & IAR-B                                & 111M                                      &       \textbf{7.80}			
				
                  &            153.3        &    \textbf{0.84}                       &    \textbf{0.39}                     & \textbf{5.77}                   & \textbf{192.5}                 & \textbf{0.85}                          & \textbf{0.42}                    \\
                              & IAR-L                                & 343M                                      & \textbf{4.35}                    & \textbf{197.2}                 & 0.81                        & \textbf{0.51}                  & \textbf{3.18}                  & \textbf{234.8}                & \textbf{0.82}                      & \textbf{0.53}                   \\
                              & IAR-XL                               & 775M                                      &         \textbf{3.15}				
                &            228.8            &          0.81                     &   \textbf{0.54}                        &         \textbf{2.52}               &   \textbf{248.1}                     &         \textbf{0.82}                     &     \textbf{0.58}                      \\
                              & IAR-XXL                              & 1.4B                         & \textbf{2.87}          &    \textbf{249.9}                   &    \textbf{0.82}               &      \textbf{0.56}                      &    \textbf{2.19}                       &       \textbf{265.6}                                        &       \textbf{0.81}                       &  0.58            \\
                              \bottomrule
\end{tabular}}
\vspace{-0.1in}
\caption{\textbf{Comparison with LlamaGen across different image tokens and model sizes.} Following LlamaGen, we only train XL and XXL version on $16\times 16$ tokens for 50 epochs.}
\label{tab:comparison with baselines}
\vspace{-0.15in}
\end{table*}

\subsection{Comparison Experiments}

{\bf Comparison with baselines.} We compare our model with the GAN-based methods~\cite{biggan,gigagan,stylegan-xl}, diffusion-based methods~\cite{adm,cdm,ldm,dit}, mask-prediction-based methods~\cite{maskgit} and autoregressive-based methods~\cite{vqgan,vit-vqgan,rq,llamagen} on ImageNet~\cite{deng2009imagenet}. The results are shown in Tab.~\ref{tab:comparison with baselines}. Our method achieves the state-of-the-art
FID (\textbf{2.19}) and IS (\textbf{362.0}). And for each model size, our IAR consistently outperforms LlamaGen, demonstrating the effectiveness of our cluster-oriented autoregressive \yrn{visual} generation strategy.


\begin{figure*}[t]
\centering
\includegraphics[width=1.0\textwidth]{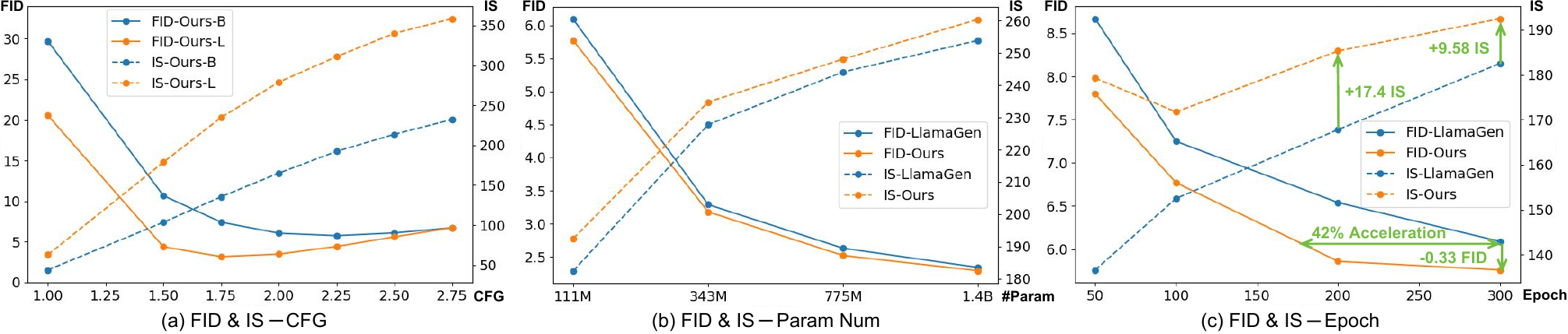}
\vspace{-0.28in}
\caption{(a) Model performance (IAR-B and IAR-L) across different CFG\yrn{s}; (b) Model performance on different parameter numbers (111M to 3B) compared to LlamaGen; and (c) Model performance on different epochs compared to LlamaGen.}
\label{fig: result-figs}
\vspace{-0.18in}
\end{figure*}

{\bf \yrn{More} comparison\yrn{s} with LlamaGen.}
We \yrn{further} compare our model with the baseline LlamaGen across different model sizes and image tokens to validate the effectiveness of our model. Specifically, we choose \yrn{four} models with model sizes ranging from 111M to 1.4B parameters. And we also conduct experiments on $16\times 16$ image tokens (corresponds to $256\times 256$ image) and $24\times 24$ image tokens (corresponds to $384\times 384$ image). We choose the model trained with 50 epochs and 300 epochs for evaluation. The comparison results are shown in Tab.~\ref{tab:comparison with baselines}. It can be seen that across different model sizes, image tokens, and training epochs, our model can always improve the baseline with a better FID and IS, validating the effectiveness and robustness of our model in LLM-based \yrn{visual} generation.

{\bf Effects of classifier-free guidance (CFG).}  IAR can employ classifier-free guidance~\cite{cfg} to generate images with higher quality. We show the performance (FID \& IS) of our model (111M and 343M parameters)
on different CFGs ranging from 1.0 to 2.75 in Fig.~\ref{fig: result-figs} (a). It can be seen that compared to the model without CFG (\emph{i.e.}, CFG=1.0), the model with CFG can largely improve the FID, which means the model can generate images with much higher \yrn{quality}. Specifically, as the CFG increases, the FID first decreases and then increases, indicating that a too-large CFG may influence the generation quality, which is consistent with LlamaGen and previous diffusion-based methods~\cite{adm,cfg}.

{\bf Effects of model size.} We conduct experiments on different model sizes, where the parameter number ranges from 111M to 1.4B. It can be seen in Tab.~\ref{tab:comparison with baselines} and Fig.~\ref{fig: result-figs} (b) that as the parameter number grows, the model shows \yrn{a} clear improvement in the performance, which aligns with the scaling law~\cite{scaling_law}, indicating that as the number \yrn{of} parameter \yrn{increases}, our IVR can be more effective.

{\bf Training efficiency.} We compare the training speed between IAR-B and LlamaGen-B (for different model sizes, please refer to \#Suppl). We evaluate the model performance on different epochs (from 50 to 300). The FID and IS results are shown in Fig.~\ref{fig: result-figs} (c), \yrn{which shows} that our 175-epoch IAR reaches almost the same FID score as the 300-epoch LlamaGen, indicating that our model can accelerate the training speed for almost $42\%$. And comparing the models trained on 300 epochs, our model 
\yrn{effectively improves the generation quality,}
validating that our model can improve the training efficiency of the LLM-based visual generation model.


\subsection{Ablation Study}
To evaluate the effectiveness of our model, we conduct several ablation studies to validate each module separately. 
We conduct the ablation studies based on IAR-B (111M parameters), and train the model for 100 epochs on ImageNet.

{\bf Ablation on codebook rearrangement and $\mathcal{L}_{CCE}$.} We first evaluate the effectiveness of the code\yrn{book} rearrangement \yrn{strategy} and the cluster-oriented cross-entropy loss $\mathcal{L}_{CCE}$
We conduct experiments on \yrn{three} ablated models: the model without both codebook rearrangement and $\mathcal{L}_{CCE}$ (original LlamaGen), the model without codebook rearrangement, and the model without $\mathcal{L}_{CCE}$. 
In contrast, our model with both codebook rearrangement and $\mathcal{L}_{CCE}$ has the best performance in fitting the ImageNet dataset.

\begin{table}[t]
\centering
\renewcommand{\arraystretch}{1.1}
\resizebox{0.47\textwidth}{!}{
\begin{tabular}{cc|cccc}
\toprule
\makecell[c]{Codebook\\Rearrangement} & $\mathcal{L}_{CCE}$&  FID$\downarrow$ & IS$\uparrow$ & Precision$\uparrow$ & Recall$\uparrow$  \\
\midrule
 & & 7.14 & 166.38 & 0.84 & 0.40 \\
\checkmark &&7.15 &163.09&0.84&0.41 \\
 &\checkmark &6.96\footnotemark[2]&168.70&0.84&0.40 \\
\checkmark &\checkmark & \textbf{6.77} & \textbf{171.73} & \textbf{0.84} & \textbf{0.42} \\
\bottomrule
\end{tabular}}
\vspace{-0.1in}
\caption{\textbf{Ablation studies on code\yrn{book} rearrangement \yrn{strategy} and the cluster-oriented cross-entropy loss $\mathbf{\mathcal{L}_{CCE}}$}.}
\vspace{-0.1in}
\label{tab:abaltion study on stage 1 and 2}
\end{table}

{\bf Ablation on cluster size.} A key hyperparameter in our model is the size of each cluster. A too-small cluster size may make it too hard for the model to predict the correct cluster index, which leads the model to perform similarly to the original LlamaGen (Note that the model with cluster size $1$ is 
the same as LlamaGen). In contrast, a too-big cluster size may cause the image embeddings \yrn{in each cluster to be} too different, such that even \yrn{if} the model predict\yrn{s} the \yrn{correct} cluster index, 
\yrn{the gap to the target image embedding is still too big.}
Therefore, it is important to choose an appropriate cluster size to balance the inner similarity in each cluster and the ease of predicting the \yrn{correct} cluster index. We conduct experiments on different cluster sizes ranging from $1$ to $512$, where the codebook size is $16384$. We compute the inner MSE \yrn{of cluster}, 
\yrn{\emph{i.e.,} the average distance from each code in a cluster to the centroid,}
and evaluate the model performance by FID, IS, Precision, and Recall in Tab.~\ref{tab:abaltion study on cluster size}. It can be seen that as the cluster size \yrn{increases}, the inner MSE distance becomes higher, indicating that the image codes in each cluster become less similar. Moreover, the model reaches the best performance when the cluster size is $128$, which happens to be the square root of the codebook size. 

\begin{table}[t]
\centering
\renewcommand{\arraystretch}{1.1}
\resizebox{0.44\textwidth}{!}{
\begin{tabular}{c|ccccc}
\toprule
Cluster Size & Inner MSE & FID$\downarrow$ & IS$\uparrow$ & Precision$\uparrow$ & Recall$\uparrow$  \\
\midrule
1 &0&7.14 & 166.38 & 0.84 & 0.40 \\
8 &0.018& 7.02	&167.09 	&0.85 	  &0.40    \\
16&0.022& 6.99  &166.39 	&0.84	      &0.42    \\
32&0.028& 6.83   &170.10 	&0.84	      &0.40    \\
64&0.034& 6.92 &171.65 	&0.85	      &0.39    \\
128&0.041& \textbf{6.77}	&\textbf{171.73}	    &0.84	      &\textbf{0.42}    \\
256&0.050& 6.81	&170.58	    &\textbf{0.85}       &0.41    \\
512&0.059& 6.94	&163.54	    &0.83	      &0.42    \\
\bottomrule
\end{tabular}}
\vspace{-0.08in}
\caption{\textbf{Ablation studies on different cluster sizes}  from $1$ to $512$, where cluster size $1$ corresponds to the official LlamaGen.}
\vspace{-0.15in}
\label{tab:abaltion study on cluster size}
\end{table}

\footnotetext[2]{\car{For details, please refer to Sec. F of \#Suppl.}}
{\bf Ablation on weight of $\mathcal{L}_{CCE}$.} Since we introduce an additional cluster-oriented cross-entropy loss $\mathcal{L}_{CCE}$ during the training process, we conduct experiments to find the influence of its corresponding weight $\lambda$. We choose $\lambda$ \yrn{ranging} from $0.1$ to $1.5$ and evaluate the model performance in Tab.~\ref{tab:abaltion study on loss weight}. It can be seen that a too-small weight will cause the model to perform similarly to the original LlamGen, while a too-big weight will make the model concern too much about the cluster index and overlook the ground-truth token index, which also makes the model perform worse. In summary, we find that when the weight $\mathbf{\lambda=1}$, the model can balance the two goals \yrn{and achieves the best performance}. 

\begin{table}[t]
\centering
\renewcommand{\arraystretch}{1.1}
\resizebox{0.33\textwidth}{!}{
\begin{tabular}{c|cccc}
\toprule
$\lambda$ & FID$\downarrow$ & IS$\uparrow$ & Precision$\uparrow$ & Recall$\uparrow$  \\
\midrule
0 & 7.14 & 166.38 & 0.84 & 0.40 \\
0.1 &7.01&167.16&0.84&0.40 \\
0.25 &7.05&160.73&0.84&0.42 \\
0.5 &6.78&171.15&0.84&0.41 \\
0.75 & 6.80& 170.06&0.84&0.41 \\
1& \textbf{6.77}	&\textbf{171.73}	&\textbf{0.84}  &\textbf{0.42}    \\
1.5 &6.87&170.60&0.84&0.41\\

\bottomrule
\end{tabular}}
\vspace{-0.05in}
\caption{\textbf{Ablation studies on the weight $\lambda$ of the cluster-oriented cross-attention loss $\mathbf{\mathcal{L}_{CCE}}$.}}
\vspace{-0.15in}
\label{tab:abaltion study on loss weight}
\end{table}

%% file: sec/5_summary.tex
\section{\yrn{Conclusion}}
\label{sec:summary}
In this paper, we analyzed the difference between the natural language and images in LLM-based visual generation. We find that similar image embedding\yrn{s} in the codebook can produce similar images. 
Based on this observation, we propose \hut{IAR, an Improved Autoregressive Visual Generation Method that effectively enhances} the training efficiency and quality of the LLM-based visual generation model. We first conduct \yrn{Codebook} Rearrangement with a balanced K-means clustering algorithm to reorder the codebook into clusters with equal sizes, where the image embeddings within each cluster share similarities. Then, we introduce a Cluster-oriented Cross-entropy Loss to enable the model to learn the target cluster, which can guarantee a good generation quality even if the model predicts a wrong image token. We validate the effectiveness of our IAR on LlamaGen, and find that IAR can largely improve the training efficiency and generation quality \yrn{across different parameter scales}. Moreover, our IAR can be \yrn{applied to various} LLM-based \yrn{visual} generation models, pointing out a new direction to improve the LLM-based \yrn{visual} generation field.

\section*{Acknowledgments}
This work was supported by Shanghai Sailing Program (22YF1420300), National Natural Science Foundation of China (No. 62302297, 72192821, 62272447, 62472282, 62472285),  Young Elite Scientists Sponsorship Program by CAST (2022QNRC001), the Fundamental Research Funds for the Central Universities (YG2023QNB17, YG2024QNA44), Beijing Natural Science Foundation (L222117).